\documentclass{article}

\PassOptionsToPackage{square, numbers, sort}{natbib}

    \usepackage[preprint]{styles/neurips_2022}

\usepackage[utf8]{inputenc} 
\usepackage[T1]{fontenc}    
\usepackage{url}            
\usepackage{booktabs}       
\usepackage{amsfonts}       
\usepackage{nicefrac}       
\usepackage{microtype}      
\usepackage{xcolor}         

\usepackage{wrapfig}
\usepackage{graphicx}
\usepackage{etoc}
\usepackage{enumitem}
\usepackage{arydshln}

\usepackage{amsmath,amsfonts,bm}

\def\eqref#1{equation~\ref{#1}}

\def\1{\bm{1}}

\DeclareMathAlphabet{\mathsfit}{\encodingdefault}{\sfdefault}{m}{sl}
\SetMathAlphabet{\mathsfit}{bold}{\encodingdefault}{\sfdefault}{bx}{n}

\DeclareMathOperator*{\argmin}{arg\,min}

\usepackage[font=small,labelfont=bf]{caption}

\newcommand{\bG}{\mathbf{G}}
\newcommand{\bV}{\mathbf{V}}

\newcommand{\bE}{\mathbf{E}}

\newcommand{\bX}{\mathbf{X}}

\usepackage{tikz}
\usetikzlibrary{decorations.pathreplacing,calligraphy}
\usetikzlibrary{arrows.meta}
\usetikzlibrary{shapes,positioning,decorations.pathmorphing}
\usepackage{standalone}
\usepackage{etoolbox} 
\usepackage{listofitems} 
\usetikzlibrary{calc}
\usetikzlibrary{decorations.markings}

\tikzset{
    ncbar angle/.initial=90,
    ncbar/.style={
        to path=(\tikztostart)
        -- ($(\tikztostart)!#1!\pgfkeysvalueof{/tikz/ncbar angle}:(\tikztotarget)$)
        -- ($(\tikztotarget)!($(\tikztostart)!#1!\pgfkeysvalueof{/tikz/ncbar angle}:(\tikztotarget)$)!\pgfkeysvalueof{/tikz/ncbar angle}:(\tikztostart)$)
        -- (\tikztotarget)
    },
    ncbar/.default=0.2cm,
}

\newcommand*{\StrikeThruDistance}{0.15cm}%
\newcommand*{\StrikeThru}{\StrikeThruDistance,\StrikeThruDistance}%
\tikzset{strike thru arrow/.style={
    decoration={markings, mark=at position 0.5 with {
        \draw [purple, thick,-] 
            ++ (-\StrikeThruDistance,-\StrikeThruDistance) 
            -- ( \StrikeThruDistance, \StrikeThruDistance);}
    },
    postaction={decorate},
}}

\tikzset{square left brace/.style={ncbar=0.2cm}}
\tikzset{square right brace/.style={ncbar=-0.2cm}}

\tikzset{>=latex} 
\colorlet{myred}{red!80!black}
\colorlet{myblue}{lightgray!80!black}
\colorlet{mygreen}{gray!60!black}
\colorlet{myorange}{blue!80!black}
\colorlet{mydarkred}{myred!40!black}
\colorlet{mydarkblue}{myblue!40!black}
\colorlet{mydarkgreen}{mygreen!40!black}
\colorlet{mydarkorange}{myorange!40!black}
\tikzstyle{node}=[very thick,circle,draw=myblue,minimum size=22,inner sep=0.5,outer sep=0.6]
\tikzstyle{connect}=[->,thick,mydarkblue,shorten >=1]
\tikzset{ 
  node 1/.style={node,mydarkblue,draw=myblue,fill=myblue!20},
  node 2/.style={node,mydarkblue,draw=myblue,fill=myblue!20},
  node 3/.style={node,mydarkgreen,draw=mygreen,fill=mygreen!25},
}
\def\nstyle{int(\lay<\Nnodlen?min(2,\lay):3)} 

\usepackage{todonotes}

\definecolor{highlightColor_blue}{RGB}{51,133,255}
\definecolor{highlightColor_yellow}{RGB}{255,255,153}

\usepackage[pagebackref=true]{hyperref} 
\hypersetup{colorlinks,linkcolor={blue!90},citecolor={blue!90},urlcolor={blue!90}} 
\renewcommand*\backref[1]{\ifx#1\relax \else (Cited on page #1) \fi}

\usepackage[capitalise]{cleveref}

\title{On the Generalization and Adaption Performance of Causal Models}

\author{%
    \parbox{\linewidth}{
        \centering 
        Nino Scherrer$^{1,2}$, 
        Anirudh Goyal $^{1}$,
        Stefan Bauer, $^{3}$,
        Yoshua Bengio $^{1}$,
        Nan Rosemary Ke $^{4}$
    }\\
    \vspace{-2mm}\\
    \parbox{\linewidth}{
        \centering 
        $^1$ Mila, Université de Montréal,
        $^2$ ETH Zurich, 
        $^3$ KTH Stockholm,
        $^4$ DeepMind
    }\\
    \vspace{-2mm}\\
    Corresponding author: \texttt{nino.scherrer@gmail.com}
  
}
\begin{document}

\maketitle

\begin{abstract}
Learning models that offer robust out-of-distribution generalization and fast adaptation is a key challenge in modern machine learning. Modelling causal structure into neural networks holds the promise to accomplish robust zero and few-shot adaptation. Recent advances in differentiable causal discovery have proposed to factorize the data generating process into a set of modules, i.e. one module for the conditional distribution of every variable where only causal parents are used as predictors. Such a modular decomposition of knowledge enables adaptation to distributions shifts by only updating a subset of parameters. In this work, we systematically study the generalization and adaption performance of such modular neural causal models by comparing it to monolithic models and structured models where the set of predictors is not constrained to causal parents. Our analysis shows that the modular neural causal models outperform other models on both zero and few-shot adaptation in low data regimes and offer robust generalization. We also found that the effects are more significant for sparser graphs as compared to denser graphs.
\end{abstract}

\section{Introduction}
Deep Learning models have demonstrated remarkable capabilities when the test distribution matches the training distribution, but their performance significantly degrades as the test distribution diverges from the training distribution \citep{lake2017building,rosenfeld2018elephant,bahdanau2018systematic,packer2018assessing,nichol2018gotta,cobbe2019quantifying, taori2020measuring, djolonga2021robustness, koh2021wilds}. However, such distribution shifts are inevitable in the real world and can occur in various settings, e.g. across hospitals in healthcare or across locations in agriculture \citep{wang2021generalizing}. This sensitivity to distribution shift inherently limits the robust and safe deployment in the wild.  At the same time, deep learning systems constructed with a multi-layered monolithic architecture tend to co-adapt different components of the network. Due to such a monolithic structure, when the distribution changes, a majority of the components of the network are likely to adapt in
response to these changes, potentially leading to poor performance on out-of-distribution samples \citep{ke2021systematic,goyal2019recurrent} and interference between subtasks or subdistributions. Endowing neural networks with the ability to capture the underlying causal structure holds the promise to accomplish  much out-of-distribution adaptation and generalization by properly factorizing
the knowledge that is stationary (causal mechanisms) from the knowledge that isn't
(the state of the random variables and interventions that change the distribution).

Given the underlying causal graph $G$, every causal mechanism represents a conditional probability distribution $p(X_i|X_{pa(i, G)})$ of a given variable $X_i$ where only causal parents $X_{pa(i)}$ are used as predictors. In such a causal framework, distribution shifts can be interpreted as interventions (i.e. perturbations) that affect certain mechanisms locally \citep{magliacane2018domain, zhang2013domain, scholkopf2021toward}. As usually not the complete environment and its structure changes at once, adapting to a distribution shift in such a framework is therefore equivalent to adapting the intervened mechanisms. 

The promising opportunities of causal models in machine learning have led to a flurry of work and accompanying advances along various research axes (e.g. causal discovery \citep{zheng2018dags, yu2019dag, bengio2019meta, ke2019learning, lachapelle2019gradient, brouillard2020differentiable, scherrer2021learning, annadani2021variational, lorch2021dibs, lippe2021efficient, geffner2022deep, ke2022learning, tigas2022interventions}, domain adaptation \citep{zhang2013domain, peters2016causal,rojas2018invariant, magliacane2018domain, bengio2019meta, le2021analysis}, robustness of neural networks \citep{zhang2020causal, kyono2020castle}, causal models in reinforcement learning (RL) \citep{de2019causal, dasgupta2019causal, nair2019causal, goyal2019recurrent, rezende2020causally, ke2021systematic}, etc. While most of these works either analyze a problem-specific objective, such as structure discovery or a success rate on a task, little attention has been paid to a \textit{systematic analysis} of the generalization and adaptation capabilities of causal models. Previous work including speed of adaptation analysis of \citet{bengio2019meta, le2021analysis} is limited to causal and anti-causal models in a bivariate setting. The work of \citet{ke2021systematic} analyzes generalization and adaptation performance of models with different inductive biases in different high-dimensional RL environments, where the underlying causal structure as well as causal variables are not given and need to be learned directly from high dimensional visual input.

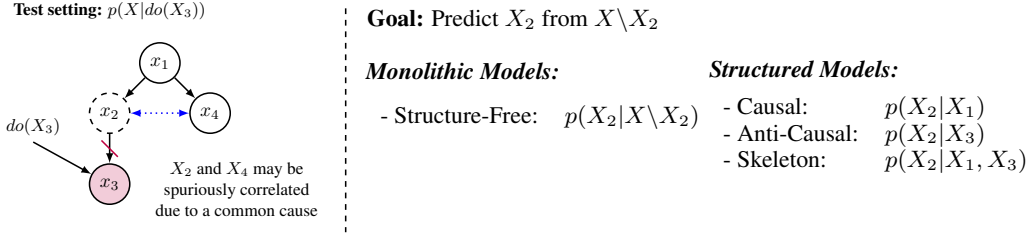
\begin{figure}[t]
    \centering
    \begin{minipage}[c]{.33\textwidth}
        \centering
            \begin{tikzpicture}[node distance={13mm}, thick, main/.style = {draw, circle}] 
        \node[main] (1) {$x_1$}; 
        \node[main, dashed] (2) [below left of=1] {$x_2$}; 
        \node[main] (4) [below right of=1] {$x_4$}; 
        \node[main, fill=purple!20] (3) [below of=2] {$x_3$};

        \draw[->] (1) -- (2); 
        \draw[->] (1) -- (4); 
        \draw[<->, dotted, blue] (2) -- (4); 
        \draw[->, black, strike thru arrow] (2) -- (3); 
        
        \node[align=center, black] (Inter) [below right = 0.5cm and -1.4cm  of 4] {\small $X_2$ and $X_4$ may be \\ \small spuriously correlated \\\small due to a  common cause};
        
        \node[align=center, black] (Int) [above = 1.2cm of 2] {\footnotesize \textbf{Test setting:} $\small p(X | do(X_3))$};

        \node[align=center, black] (Int) [above left= 0.5cm and 0.5cm  of 3] {\footnotesize $\small do(X_3)$};        
        \draw[->] (Int.south) -- (3);
        
        \draw [dashed] (3.4,1) -- (3.4,-3.1);

    \end{tikzpicture} \\
    \end{minipage}
    \hspace{1mm}
    \begin{minipage}{.32\textwidth}\vspace{0pt}
        \small
        \vspace{-4mm}
        \textbf{Goal:} Predict $X_2$ from $X{\setminus}X_2$ \hspace{1mm}\\\quad\\
        \textbf{\textit{Monolithic Models:}} 
         $$
            \scriptstyle
            \begin{array}{ll}%
                \text{- Structure-Free:}  & p(X_2|X {\setminus} X_2) 
            \end{array}
        $$
        \vspace{0mm}\\
    \end{minipage}
    \begin{minipage}{.32\textwidth}\vspace{0pt}
        \small
        \vspace{-4mm}
        \quad 
        \hspace{1mm}\\\quad\\
        \textbf{\textit{Structured Models:}} 
        $$
            \scriptstyle
            \begin{array}{ll}%
                \text{- Causal:}          &  p(X_2|X_1)  \hspace{4mm}\\
                \text{- Anti-Causal:}     &  p(X_2|X_3)  \hspace{4mm}\\
                \text{- Skeleton:}        &  p(X_2|X_1, X_3)  \hspace{2mm}\\
            \end{array}
        $$
    \end{minipage}
    \caption{\textbf{Predictions under Distribution Shift.} A hard intervention $do(X_3)$ leads to a distribution shift which breaks the dependence between $X_2$ and $X_3$. \textit{Monolithic models} and \textit{structured models} that allow anti-causal predictors may still rely on $X_3$ and lead to erroneous predictions. In contrast, a structured model that only relies on causal predictors would not be affected by such a distribution shift and rely on the stable predictor $X_1$.}
    \label{fig:figure1}
    \vspace{-1.5\baselineskip}
\end{figure}

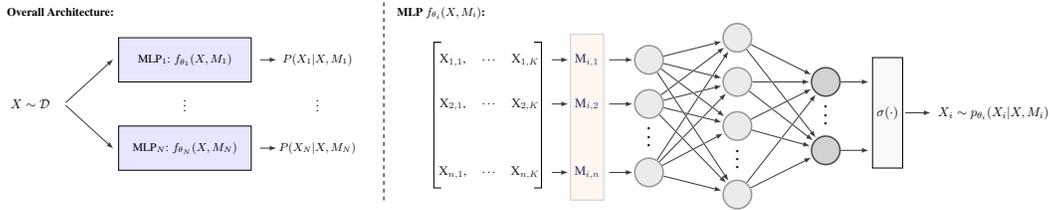
\begin{figure}[t]
    \centering
\begin{tikzpicture}[x=2.4cm,y=1.2cm]
  \readlist\Nnod{3,4,2} 
  \readlist\Nstr{n,m,k} 
  \readlist\Cstr{x,h^{(\prev)},y} 
  \def\yshift{0.55} 
  \def\M{$M$}
  \def\X{$X$}
  
  \foreachitem \N \in \Nnod{
    \def\lay{\Ncnt} 
    \pgfmathsetmacro\prev{int(\Ncnt-1)} 
    \foreach \i [evaluate={\c=int(\i==\N); \y=\N/2-\i-\c*\yshift;
                 \x=\lay; \n=\nstyle;
                 \index=(\i<\N?int(\i):"\Nstr[\n]");}] in {1,...,\N}{ 
      \node[node \n] (N\lay-\i) at (\x,\y) {};
      
      \ifnumcomp{\lay}{>}{1}{ 
        \foreach \j in {1,...,\Nnod[\prev]}{ 
          \draw[white,line width=1.2,shorten >=1] (N\prev-\j) -- (N\lay-\i);
          \draw[connect] (N\prev-\j) -- (N\lay-\i);
        }
        \ifnum \lay=\Nnodlen
          \draw[connect] (N\lay-\i) --++ (0.5,0); 
        \fi
      }{
        \node[align=center,mydarkgreen] at (-1.2,\y) {$\strut\X_{\index,1}$, };
        \node[align=center,mydarkgreen] at (-0.8,\y) {\dots };
        \node[align=center,mydarkgreen] at (-0.4,\y) {$\strut\X_{\index,K}$};
        \draw[connect] (-0.1,\y) -- (0.1,\y);
        \node[align=center,mydarkorange] at (0.3,\y) {$\strut\M_{i,\index}$};
        \draw[connect] (0.55,\y) -- (N\lay-\i); 
      }
      
    }
    \path (N\lay-\N) --++ (0,1+\yshift) node[midway,scale=1.6] {$\vdots$}; 
  }
  

  \node[draw, rectangle, fill=lightgray!10, minimum width=0.8cm, minimum height=3cm] at (3.7 ,-0.7) {$\sigma(\cdot)$};
  \draw[connect] (3.9 ,-0.7) --++ (0.3,0.0);
  
 \draw [black] (-1.35,-2.4) to [square left brace] (-1.35,0.9);
 \draw [black] (-0.3,-2.4) to [square right brace] (-0.3,0.9);
 \node[ minimum width=0.8cm, minimum height=3cm] at (4.9 ,-0.7) {$X_i \sim p_{\theta_i}(X_i |X, M_i)$};
 
 \node[draw, black, rectangle, fill=orange!30, minimum width=0.9cm, minimum height=4.5cm, opacity=.2] at (0.3 ,-0.8) {};
 
 \draw[draw=none] (-1.5, 1.0) rectangle (-0.8, 1.5) node[anchor=base east] {\small \textbf{MLP $f_{\theta_i}(X, M_i)$:}};
  \draw[draw=none] (-6.0, 1.0) rectangle (-5, 1.5) node[anchor=base east] {\small \textbf{Overall Architecture:}};
  
 \draw[dashed, line width=0.3mm, bend left=50, thick, black] (-2, 1.8) -- (-2,-2.7);
    
 \draw[draw=none,align=center,fill=white] (-6,-1) rectangle (-6, 0) node[pos=.5] {$X \sim \mathcal{D}$};
 
 \draw[connect] (-5.6,-0.47) --++ (0.55,1.0);
 \draw[connect] (-5.6,-0.53) --++ (0.55,-1.0);
 
 \draw[connect] (-3.4,0.5) --++ (0.2, 0);
 \draw[connect] (-3.4,-1.5) --++ (0.2, 0);
 
 \draw[black,rounded corners=0, align=center,fill=blue!10] (-5,1) rectangle (-3.5, 0) node[pos=.5] {\footnotesize MLP$_1$: $f_{\theta_1}(X,M_1)$};
 \draw[draw=none, align=center,fill=white] (-2.5,1) rectangle (-3, 0) node[pos=.5] {\footnotesize $P(X_1|X, M_1)$};
 
 \draw[draw=none,align=center] (-5,-1) rectangle (-3.5, 0.1) node[pos=.5] {\vdots};
 \draw[draw=none,align=center] (-2,-1) rectangle (-3.5, 0.1) node[pos=.5] {\vdots};
    
 \draw[black,rounded corners=0, align=center,fill=blue!10] (-5 ,-2) rectangle (-3.5, -1) node[pos=.5] {\footnotesize MLP$_N$: $f_{\theta_N}(X,M_N)$};
 \draw[draw=none, align=center,fill=white] (-2.5,-2) rectangle (-3, -1) node[pos=.5] {\footnotesize $P(X_N|X, M_N)$};
  
\end{tikzpicture}
    \caption{\textbf{Architectural Setup:} We employ the same model architecture consisting of a stack of MLPs across all considered models. Every MLP $f_{\theta_i}(X, M_i)$ consisting of a masking layer $M_i$ and one hidden layer, takes a data sample $X \sim \mathcal{D}$ as input, applies a input mask $M_i$ to the input and outputs the logits of a categorical distribution on its output layer. The logits are finally normalized through a softmax-activation function $\sigma(\cdot)$ which results in the CPD $p(X_i|X, M_i)$. Between the considered models, we vary the learning objective to learn the parameters $\theta$ and the inputs masks $M$. While \textit{monolithic models} and \textit{structured models} with expert knowledge have a fixed mask $M$, the causal discovery model optimizes $M$ using an additional set of parameters.}
    \label{fig:architectural_setup}
\end{figure}

In this work, we  \textit{systematically} investigate zero  and few-shot adaptation capabilities of \textit{monolithic models} and \textit{structured models} where causal variables are explicitly given. As an evaluation setting, we consider the task of predicting missing values (e.g. given all the other variables of the sample) under unseen distribution shifts (see \Cref{fig:figure1}). In order to investigate the effect of the different inductive biases, we employ the same model architecture (i.e. one MLP per conditional distribution) across all considered models (see \Cref{fig:architectural_setup}). Hence, all models have the same expressive abilities and only vary in their training objectives and pre-existing domain knowledge.   Within the class of \textit{structured models}, we distinguish between expert knowledge models where we provide certain structure upfront of training (e.g. causal graph, anti-causal graph) and models where causal structure is learned from data. We train monolithic models with different training objectives including a pseudo-likelihood objective as well as a meta-learning objective which explicitly optimizes the parameters of the monolithic models to adapt quickly to changes in distribution, hence confounding many different problems. This setup allows us to uncover generalization and adaptation discrepancies between different models and analyze if and where models are prone to fail.
 
\textbf{Contributions.} (i) We show that generalization capabilities of different models vary significantly with the amount of available training samples. (ii) We demonstrate that \textit{structured models} significantly outperform \textit{monolithic models} in low-data regimes. (iii) We show that a general evaluation metric is prone to drawing erroneous conclusions with respect to robustness and show how a general evaluation metric can be dissected into refined metrics to investigate if and how specific models fail. (iv) We show that non-causal models can fail drastically in settings where the underlying causal structure is sparse. (v) We evaluate few-shot adaptation in various settings and show that causal models are the fastest and most robust to adapt. (vi) We show how models adapt in parameter space and relate this to the speed of adaptation and robustness. (vii) We propose and investigate a new adaptation objective for causal models which enables an efficient adaptation in low training and adaptation-data regimes.

\section{Background}
In our work, we consider setting that high-level causal variables are observed and given (i.e. they do not need to be inferred from high-dimensional input). We limit the number of variables $N \in \lbrace 10, 20 \rbrace$ where causal variables $X = \lbrace X_1, \dots, X_n \rbrace$ are directly observed and assume no hidden confounding variables (i.e. causal sufficiency). We generate synthetic observational and interventional data $\mathcal{D} = (\mathcal{D}_{obs}, \mathcal{D}_{int})$ on causal acyclic graphs and fit different models to data in order to learn conditional probability distributions $p(X_i|\cdot)$ for all variables $X_i$. During test time (with and without adaptation), we predict all variables $X_j$ for $j \in \lbrace 1, \dots, N  \rbrace$ of unseen interventional distributions  $p(X | do(X_k))$ using all other variables $X_i$ ($i\not=j$) of the sample.

\textbf{Causal Graph.} A causal graph is commonly represented by a directed acyclic graph (DAG) $\bG = (\mathbf{V},\mathbf{E})$ with $|\bV|=N$ and $|\bE|=M$. Such a graph is defined over finite set of vertices $\bV$ associated with a finite set of random variables (or observables) $\bX = \lbrace X_1, \ldots, X_N\rbrace$, where directed edges in the causal graph $\bG$ point from causes to effects. For convenience, the set of $X_i$'s parents in $\bG$ is usually denoted as $X_{pa(i,\bG)}$ and the set of $X_i$'s children in $\bG$ by $X_{ch(i,\bG)}$.

\textbf{Adjacency Matrix.} The connectivity between vertices $\bV$ in a graph is commonly represented by an \emph{adjacency matrix} $\mathbf{A} \in \lbrace0,1\rbrace^{N \times N}$ such that $\mathbf{A}_{i,j}=1$ if node $j$ is a parent of node $i$. 

\textbf{Structural Causal Model (SCM).} An SCM \citep{pearl1995causal, peters2017elements}, also known as structural equation model (SEM), is defined by a causal graph $\bG$ over a set of random variables (or observables) $\bX = \lbrace X_1, \ldots, X_N\rbrace$ and a set of associated structural equations. The structural equations express the functional relationships among the causal variables through functions $f_i$ and jointly independent noise variables $U_i$ as $
    X_i \:= f_i(X_{pa(i, G)}, U_i) \forall i \in \{1, \ldots N\}$. 
The noise variables $U_i$ ensure that the set of structural equations can represent general conditional probability distributions $P(X_i|X_{pa(i,G)})$. The joint distribution entailed by the variables $\bX = \lbrace X_1, \ldots, X_N\rbrace$ can be factorized such that each variable is conditionally independent of other variables given its parents in the graph $\bG$: 
\vspace{-1mm}
\begin{equation}
    \small
    P(X_1, \ldots, X_N) = \prod_{i=1}^{N} P(X_i|X_{pa(i, \bG)}) 
\end{equation}
\vspace{-1mm}
In the causality literature, this factorization is also known as the causal factorization \citep{scholkopf2021toward}.

\textbf{Independent Causal Mechanisms (ICM) Principle.} The causal factorization can be seen as a composition of \emph{independent} causal mechanisms (ICM)~\citep{scholkopf2021toward}. The ICM principle tells us that changing one mechanism $P(X_i|X_{pa(i)})$ does not change any of the other mechanisms $P(X_j|X_{pa(j)})$ $(i\not = j)$ \citep{parascandolo2018learning, bengio2019meta, scholkopf2021toward}. This led to the Sparse Mechanisms Shift hypothesis, i.e. that small distribution changes tend to manifest themselves in a sparse or local way in the causal factorization \citep{scholkopf2021toward}.

\section{Related Work}
\textbf{Differentiable Causal Discovery.} Recent advances in differentiable causal discovery focused on building new algorithms for causal discovery from observational data \citep{zheng2018dags, yu2019dag, lachapelle2019gradient, annadani2021variational, lorch2021dibs, cundy2021bcd, geffner2022deep, ke2022learning} or fused data (observational and interventional data) \citep{ bengio2019meta, ke2019learning, brouillard2020differentiable, lippe2021efficient} using advances in deep learning. Such methods are primarily concerned to identify the underlying causal structure from data, and not evaluate the zero and few-shot capabilities of the learned models.

\textbf{Speed of Adaptation} While \citet{bengio2019meta,ke2019learning} included an analysis for adaptation or generalization speed of causal models compared to monolithic models, these analyses are focused on a specific setting, such as a specific number of variables. The work of \citet{le2021analysis} analyzes the speed of adaption of causal and anti-causal models, however, the analysis is only limited to the bivariate settings. \citet{scholkopf2021toward} discussed the generalization and adaption performance of causal models against monolithic models in a high-dimensional setting, however, no experimental analysis is included. \citet{ke2021systematic} proposed a novel suite of RL enviromments and tasks for analyzing causal discovery in a high-dimensional RL setting, and the work  analyzed  generalization and adaption performance of models with different inductive biases. In our work, we perform a systematic analysis of generalization and adaption performance of causal models against monolithic models, as well as models that are explicitly optimized using meta-learning objectives such as MAML~\citep{finn2017model} on settings where the causal variables are explicitly given.

\textbf{Domain Adaptation.} Multiple approaches have been proposed that exploit the causal structure of the data generating process in order to address the problem of domain adaptation \citep{zhang2013domain, peters2016causal, bareinboim2016causal, rojas2018invariant, magliacane2018domain}. While these works analyze specific instances of domain adaptation problems with varying assumptions, our work is concerned to investigate zero- and few-shot adaptation abilities of various \textit{monolithic} and \textit{structured models}.

\textbf{Improving Robustness through Causal Structure.} \citet{zhang2020causal} showed a connection between the vulnerability / robustness of neural neural networks and their lack of causal reasoning.  \citet{kyono2020castle} showed that learning causal structure as an auxiliary tasks improves the in-distribution generalization capabilities of overparameterized feed-forward neural networks. However, the work only investigates out-of-sample generalization within the same distribution, and does not consider out-of-distribution settings.

\textbf{Transfer through Modular Knowledge Decomposition.} Recent work has shown that architectural inductive biases which promote modular decomposition of knowledge  can provide a useful basis for transfer of knowledge from one task to another task \cite{alet2018modular, chen2020modular, madan2021fast}. Such architectures employ a meta-learning approach to update different subset of parameters of the network over different timescales and show such an approach  leads to improvements in sample efficiency as compared to training all the parameters at once \cite{madan2021fast}. Such methods learn directly from low dimensional pixel data and don't explicitly learn causal variables.  

\vspace{-2mm}
\section{Data Generation}
\vspace{-2mm}
In order to systematically investigate the effect of different training objectives on generalization and adaptation performance, we employ a synthetic data generation setup. We generate observational and interventional data $\mathcal{D} = (\mathcal{D}_{obs},\mathcal{D}_{int})$  of discrete and non-linear nature governed by causal graphs. 

\textbf{Graph Generation.} We distinguish between \emph{structured} and \emph{random} graphs in order to represent a wide diversity of possible graphs. For structured graphs, we follow the setup of \citep{ke2019learning} and generated various DAGs with acyclic and cyclic skeletons. In order to generate random graphs, we follow the Erdós–Rényi (\texttt{ER}) model with varying edge densities (i.e. \texttt{ER-1}, \texttt{ER-2} and \texttt{ER-3}) as in \citep{scherrer2021learning}. 

\textbf{Conditional Probabilty Distributions (CPD)}  
In order to generate discrete, observational data $\mathcal{D}_{obs}$ given a  causal DAG $G$, we perform ancestral sampling based on the topological order of the causal DAG $G$ as proposed in \citep{ke2019learning}. 
Similar to \citep{ke2019learning,lippe2021efficient}, we model the CPDs $p(X_i|X_{pa(i,G)})$ using randomly initalized one-hidden layer MLPs (weights orthogonally in the range $[-2.5, 2.5]$ and biases uniformly in the range $[-1.1, 1.1]$) with a hidden dimensionality of $48$ where all inputs except the parents $X_{pa(i,G)}$ are masked to $0$ during the sampling process. We perform point interventions on a single node modelled by an Uniform distribution $U[1, K]$, where $K$ is the number of possible categorical assignments. 

\vspace{-2mm}
\section{Model Architecture and Training Setup}
\vspace{-1mm}

\subsection{Model Architectures}
\vspace{-1mm}
\label{sec:architecture}
In order to disentangle architectural effects from the training objectives, we employ the same architecture across all evaluated models. We follow the setup of \citep{ke2019learning, scherrer2021learning} and choose a stack of $N$ independent MLPs for a setting with $N$ observed variables. Hence for every variable $X_i$, there exists an MLP parametrized by $\theta_i$ that represents the conditional probability distribution (CPD) $p(X_i|X, M_i)$ where $M_i\in\lbrace 0,1 \rbrace^{N}$ denotes a input mask. Specifically, every MLP has an input layer of size $N \times K$, one hidden layer of $64$ neurons with Leaky ReLU activations of slope $0.1$ and a linear output layer of size $K$. The output layer represents the unnormalized log-probabilities of each possible category that are finally mapped to valid CPDs through a softmax activation function $\sigma(\cdot)$.

\subsection{Training Objectives}
\label{sec:model_introduction}

In our study, we evaluate six different learning paradigms on a fixed architecture consisting of a stack of $N$ MLPs (see \cref{sec:architecture} for a detailed description) with the goal to learn the underlying generative mechanism $p(X_i|\cdot)$. To this end, we consider two techniques to learn a \textit{monolithic model} (i.e. \texttt{Pseudo-LL} and \texttt{MAML}) and four different techniques to learn a \textit{structured model}  (i.e. \texttt{EXP-Causal}, \texttt{EXP-AntiCausal}, \texttt{EXP-Skeleton} and \texttt{L-Causal}). Within the \textit{structured models}, we construct three models using expert knowledge (\texttt{EXP}) and a model where the causal structure is learned from data (i.e. \texttt{L-Causal})  without any supervision. The exact training objectives are described in the following paragraphs:

\subsubsection{Monolithic Models}
    \textbf{Pseudo-Loglikelihood (\texttt{Pseudo-LL}):} As a simple monolithic model, we train a model using maximum likelihood on an unconstrained input mask $M$ on observational data $\mathcal{D}_{obs}$. We minimize:
    \begin{equation}
        \label{eq:MLE}
        \small
        \theta_i^{*} = \argmin_{\theta_i} \mathop{\mathbb{E}}_{X\sim\mathcal{D}_{obs}} [-log(f(X, M_i;\theta_i))]
    \end{equation}
    for every MLP$_i$ independently, where $M_i$ denotes the $i$'th column of the input mask and $\theta_i$ the corresponding MLP parameters. $M$ is a unconstrained input mask with ones everywhere except zeroes on the diagonal. This prevents from learning an identity mapping $X_i = X_i$. Hence, the \texttt{Pseudo-LL} model learn CPD's of the form  $p(X_i | X\setminus X_i)$ on data from $\mathcal{D}_{obs}$.

    
    \textbf{Model-Agnostic Meta-Learning (\texttt{MAML}):} Motivated by the adaptation capabilities of meta-learned models, we use the model-agnostic meta learning algorithm (MAML) \citep{finn2017model} in order to train a monolithic model on a variety of interventional distributions  $\lbrace \mathcal{D}_{int(l)} \rbrace_{l=1}^{L}$. To this end, we employ the following meta-optimization formulation: 
    \begin{equation}
        \small
        \begin{split}
        \small
        \theta_i^{*} & =\argmin _{\theta_i} \sum_{l \sim p(Int)} \mathop{\mathbb{E}}_{X\sim\mathcal{D}_{int(l)}} \big[-log\big(f(X, M_i;\hat\theta_{i,l})\big)\big] \\
        \theta_i^{*}  & =\argmin _{\theta_i} \sum_{l \sim p(Int)} \mathop{\mathbb{E}}_{X\sim\mathcal{D}_{int(l)}} \bigg[-log \bigg(f \big(X, M_i; \underbrace{\theta_{i} - \alpha \nabla_{\theta_{i}} \mathop{\mathbb{E}}_{X\sim\mathcal{D}_{int(l)}} \big[-log(f(X, M_i; \theta_i \big]}_{:= \hat\theta_{i,l}} \big) \bigg)\bigg]
        \end{split}
    \end{equation}
     where $p(Int) = \mathcal{U}[1, \dots, L]$ denotes a uniform distribution over the available interventional distributions, $M_i$ is the input mask, $\alpha$ is the step-size parameter of the inner update and $\hat\theta_i,l$ represents the updated model parameters. The above meta-optimization objective is optimized using Adam \citep{kingma2014adam} and the inner updated is done using stochastic gradient descent (SGD). Note that the meta-optimization is performed over the model parameters $\theta_i$, whereas the objective is computed using the updated model parameters $\hat\theta_i$. For the input mask $M_i$, we follow same setup as for \texttt{Pseudo-LL}. Hence, the \texttt{MAML} models learns CPD's of the form  $p(X_i | X\setminus X_i)$ on different interventional distributions $\mathcal{D}_{int(l)}\sim \mathcal{D}_{int}$. In our experiment, we rely on the first-order approximation of MAML \citep{nichol2018first}.
     
    \subsubsection{Structured Models}  
     \textbf{Learning with Expert Knowledge} (\texttt{EXP-Causal}, \texttt{EXP-AntiCausal} and \texttt{EXP-Skeleton})\textbf{:} Given the adjacency matrix $A$ of the ground-truth causal structure $G$, the anti-causal adjacency matrix (i.e. the transpose of the causal adjacency) or the adjacency matrix of the undirected skeleton, we inject the the provided expert knowledge by setting the input mask to $M=A$ and train the models \texttt{EXP-Causal}, \texttt{EXP-AntiCausal} and \texttt{EXP-Skeleton} using maximum likelihood training on observational data $\mathcal{D}_{obs}$. To this end, we minimize \Cref{eq:MLE} for every MLP independently. Hence, \texttt{EXP-Causal} learns CPD's of the form $p(X_i | X_{pa(i, G)})$ (i.e. only causal predictors), \texttt{EXP-AntiCausal} learns CPD's of the form $p(X_i | X_{ch(i, G)})$ (i.e.  only anti-causal predictors) and \texttt{EXP-Skeleton} learns CPD's of the form $p(X_i | X_{pa(i, G)}), X_{ch(i, G)})$ (i.e. causal and anti-causal predictors).

    \textbf{Learning Causal Structure (\texttt{L-Causal}):} We use a  causal discovery framework to learn a structural causal model (SCM) from data. To this end, we introduce an additional set of parameters $\gamma = (u, v)$ with $u \in \mathbb{R}^{N \times N}$ and $v \in \mathbb{R}^{N \times N}$ which define a continuous relaxation of an adjacency matrix $\gamma = \sigma(u) \cdot \sigma(v)$. Such a soft-adjacency matrix can be conveniently used to sample input masks $M$. In order to train the parameters $\theta$ of the MLPs  and the learnable input mask $\gamma$, we rely on a optimization formulation as in \citep{ke2019learning, scherrer2021learning} using two alternating phases of optimization. These are performed until convergence in an iterative manner. Under freezed mask parameters $\gamma$, we train during phase 1 (called "CPD Fitting") the parameters $\theta_i$ of each MLP on observational data $\mathcal{D}_{obs}$ using a similar maximum likelihood objective as in \Cref{eq:MLE}: 
    \begin{equation}
        \small
        \theta_i^{*} = \argmin_{\theta_i} \mathop{\mathbb{E}}_{X\sim\mathcal{D}_{obs}} \mathop{\mathbb{E}}_{M \sim \sigma(\gamma)}[-log(f(X, M_i;\theta_i))]
    \end{equation}
    where we sample a set of input masks $M$ from $\gamma$ instead of relying on a fixed mask $M$. During phase 2 (called "Mask Fitting"), we freeze the previously trained MLP parameters $\theta$ and optimize the mask parameters $\gamma$ using different sets of interventional data $\mathcal{D}_{int(l)}\sim \mathcal{D}_{int}$ using the optimization objective as proposed in \citep{lippe2021efficient}. For the exact optimization objective and further implementation details, we refer to the appendix \S \ref{appendix:implementation_modelArchitecture}. In contrast to the \textit{structured models} with injected expert knowledge (i.e. fixed masks), we model the causal structure as learnable soft adjacency matrix. This continuous relaxation is then used to sample different input masks for distribution fitting. After convergence, the model \texttt{L-Causal} represents $p(X_i | X_{pa(i, G)})$ which is adherent to the "learned" causal structure $G$. 
\subsection{Adaptation Techniques}
\label{sec:adaptation_techniques}
During test time, we aim to adapt  pretrained models to interventional distributions that were not presented during training. Given a set of adaptation samples $\mathcal{D}_{int}^{A}$, we consider:

\begin{itemize}[topsep=2pt,itemsep=0pt,leftmargin=15pt]
    \item \textbf{Unconstrained Adaptation.} Finetune all MLPs using $\mathcal{D}_{int}^{A}$ for multiple gradient steps.
    \item \textbf{Sparse Adaptation.} We only finetune the module that was affected by the intervention. The affected module is either known or predicted in the setting of unknown interventions.
    \item \textbf{Regularized Adaptation.} By relaxing the Sparse Mechanisms hypothesis, we aim to compute adaptation weights that control the magnitude of adaptation (lower adaptation where knowledge can be directly reused, more adaptation where knowledge has changed). To this end, we compute the NLL on the adaptation data for every variable $X_i$ and store it as score $s_i$, assessing the fit of each MLP given the new transfer data. We use this score to compute the weight of adaptation of a certain mechanism by employing a temperature-scaled softmax over these scores  $[s_1,\dots,s_N]$. The temperature $t$ allows to control the magnitude of co-adaptation and interpolate between unconstrained adaptation ($t=\infty$) and sparse adaptation ($t=0$). 
\end{itemize}


\vspace{-1mm}
\section{Analysis of Generalization Performance}
\label{sec:generalization}
\vspace{-1mm}
We start by analyzing  OOD generalization (zero-shot) performance for the different models discussed in \Cref{sec:model_introduction}. In particular, our experiments seek to answer the following questions: (a) How different models generalize under different circumstances. (b) Analyzing how different parts of our models contribute to the failure or success for OOD generalization.

\textbf{Implementation details.}
We keep the training and evaluation setup between different models as similar as possible. All models have the same number of parameters to represent the CPDs and are trained for the same number of steps, we train models on the same range of learning rates and pick the best performing one for each model individually. All experiments are run with $10$ random seeds, we report the mean and standard deviation of the results. In particular, we use the Adam optimizer \citep{kingma2014adam} across all models and have evaluated learning rates from the set $\lbrace 1$e$-2, 1$e$-3, 1$e$-4 \rbrace$, weight decay from the set $\lbrace 1$e$-4, 1$e$-5, 0 \rbrace$ and train all models for $1000$ iterations, except the \texttt{L-Causal} model which was trained for multiple rounds with iterative optimization. For the detailed setup with all model-specific hyperparameters, we refer to appendix \S \ref{appendix:implementation_hyperparameters}.

\textbf{Performance Bounds on Causal Models.} To assess the performance of the causal models, we compute two upper bounds on the performance of the causal model by accessing the data-generating causal model. (a) \texttt{Bound-ZeroShot} shows the maximal zero-shot adaptation performance and (b) \texttt{Bound-Adaptation} show the maximal adaptation performance on the transfer distribution. Note that the causal model $\texttt{EXP-Causal}$ that relies on expert knowledge should naturally attain this bound faster than the model \texttt{L-Causal} that learns the causal structure from data.

\vspace{-2mm}
\begin{figure}[t!]
    \vspace{-0.5\baselineskip}
    \centering
    \includegraphics[width=1.0\linewidth]{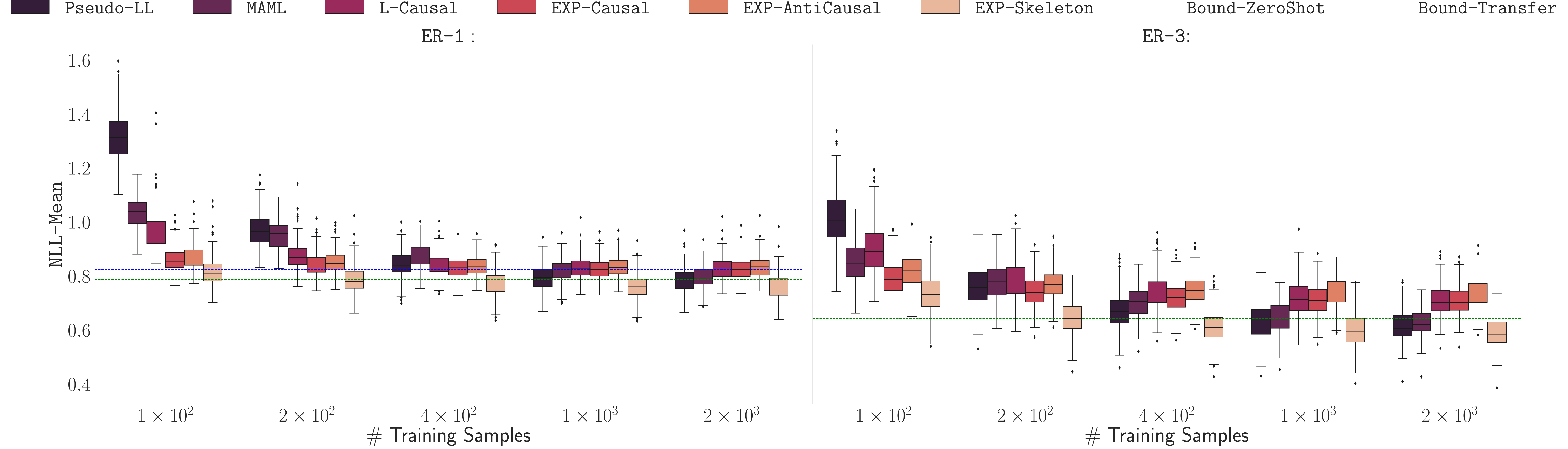}
    \caption{\textbf{OOD Generalization with Varying Amounts of Training Data.} We report \texttt{NLL-Mean} over a set of unseen interventions across \texttt{ER}-graphs of varying density ($N=20$ and 10 graphs per setting) and training sets of increasing size (up to 2k samples). \textit{Structured models} outperform \textit{monolithic models} consistently on low data regimes on the \texttt{NLL-mean} metric. \texttt{EXP-Skeleton} outperforms all other models on all settings with respect to this metric. The causal models attain \texttt{Bound-ZeroShot} (green) as expected with sufficient amount of samples.
    } 
    \label{fig:results_zeroshot_trainingData}
    \vspace{-1\baselineskip}
\end{figure}
\subsection{Generalization performance.}
\label{sec:analysis_generalization_general}
We analyze the generalization performance of different models under different settings. There are three aspects of the settings that we consider. First is the amount of training data $\mathcal{D}$, then we look at the  density of the underlying SEM that generated the data; at last, we look at the size of the graph. To be specific, we vary the amount of training data between  $10^2$ and $10^4$,  the graph density of the underlying SEM between ER-1 and ER-3 and the size of the graph between $10$ and $20$.

The training data $\mathcal{D}^T$ consist of both observational $\mathcal{D}^T_{obs}$ and interventional data $\mathcal{D}^T_{int}$. We keep the number of observational data $\mathcal{D}^T_{obs}$ the same as the number of  interventional data $\mathcal{D}^T_{int}$ in training. Test data $\mathcal{D}^t$ is kept fixed across  experiments.  We report the 
\texttt{NLL-Mean} of the test data on the model, this is  average NLL scores across all variables (including the intervened-on variable). 

\textbf{Summary.} Results for comparisons between different amount of data and graph sparsity are found in Figure \ref{fig:results_zeroshot_trainingData}, results for comparisons between data with different number of nodes are in Figure \ref{fig:results_zeroshot_overNumberOfVariables}. Refer to $\S \ref{appendix:dissection_all}$ in the appendix for a complete set of results. We found  that \emph{causal models} outperform \emph{monolithic models} (\texttt{Pseudo-LL} and \texttt{MAML}) when the amount of training data is low (Figure \ref{fig:results_zeroshot_trainingData}). We also found that the performance gap widens as the density of the graphs decreases (Figure \ref{fig:results_zeroshot_trainingData}). The performance gap also widens as the size of the SEM increases (Figure \ref{fig:results_zeroshot_overNumberOfVariables}). These results suggest that models with the correct structure (such as causal models) generalize better compared to models with no structure, especially when trained on a small amount of data coming from sparse and large graphs. Furthermore, \texttt{EXP-Causal} and \texttt{EXP-Skeleton} models outperforms \texttt{EXP-AntiCausal} models under all settings. This suggests that having the correct structure is important and having the wrong structure (i.e. no causal predictors) can hurt performance. 

We observe slower convergence of \texttt{causal} models to \texttt{Bound-ZeroShot} on dense graphs than on sparse graph, as the identification of the causal structure is more challenging in such settings \citep{ke2019learning, scherrer2021learning}. An interesting observation is that \texttt{EXP-Skeleton} models can outperform \texttt{EXP-Causal} models, this is because that when the intervention is not on the children of the predicted node, then \texttt{EXP-Skeleton} models can use the value of the child and the the value of the parents to predict the value of the node, whereas, the \texttt{EXP-Causal} models only uses the parents to predict the value of the node. However, the   \texttt{EXP-Skeleton} models will fail catastrophically when the intervention is on the children of the predicted node, we will see more analysis about this in Section \ref{sec:dissection}.

\vspace{-1mm}
\subsection{Dissecting generalization performance}
\label{sec:dissection}

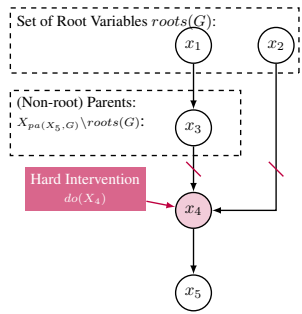
\begin{wrapfigure}{r}{0.3\textwidth}
    \vspace{-3mm}
    \centering
        \begin{tikzpicture}[node distance={17mm}, thick, main/.style = {draw, circle}] 
        \node[main] (1) {$x_1$}; 
        \node[main] (3) [below of=1] {$x_3$}; 
        \node[main] (2) [right of=1] {$x_2$}; 
        \node[main, fill=purple!20] (4) [below of=3] {$x_4$}; 
        \node[main] (5) [below of=4] {$x_5$}; 
        \draw[->] (1) -- (3); 
        \draw[->, black, strike thru arrow] (3) -- (4); 
        \draw[->, black, strike thru arrow] (2) -- (+1.7, -3.4) -- (4);
        \draw[->] (4) -- (5);
        
        \node[align=center, white, fill=purple!60] (Int) [above left = -0.3cm and 0.7 cm of 4] {\footnotesize Hard Intervention \\ $\scriptstyle do(X_4)$};
        
        \draw[->, purple] (Int) -- (4);
        
        \draw[thick, dashed] ($(2.south east)+(0.2,-0.3)$) rectangle ($(1.north west)+(-3.5,0.5)$) node[below right] {\footnotesize Set of Root Variables $roots(G)$:};
        \draw[thick, dashed, align=left] ($(3.south east)+(0.7,-0.3)$) rectangle ($(3.north west)+(-3.5,0.5)$) node[below right] {\footnotesize (Non-root) Parents: \\ $\scriptstyle X_{pa(X_5, G)} \setminus roots(G)$:};
    \end{tikzpicture} 
    \label{fig:dissect_fig}
    \caption{\textbf{Dissection Illustration.} Based on the topology of  a  causal graph, we dissect the nodes into subcategories.}
    \label{fig:dissect_fig}
    \vspace{-2\baselineskip}
\end{wrapfigure}
The analysis in \Cref{sec:analysis_generalization_general} reports an average evaluation metric across all nodes, which may not help us to understand the model's performance in detail. In this set of experiments, we aim to better understand the model's performance by  dissecting the \texttt{NLL-Mean} metric into sub-metrics. This could help to systematically identify if and where models are prune to failure. Such an analysis helps us to assess the the individual model robustness to distribution shifts and uncover failure settings which are hidden in the \texttt{NLL-Mean} metric. To this end, we consider a data regime ($1$k training samples) where all models perform similar with respect to the general \texttt{NLL-Mean} score. 


We dissect $\texttt{NLL-Mean}$ systematically into: (a) \texttt{NLL-Intervention:} NLL on intervened variable $X_i$, (b) \texttt{NLL-Root:} Mean-NLL on root variables in $G$, (c) \texttt{NLL-Parents:} Mean-NLL on parent variables $X_{pa(i,G)}$ for a intervention on $X_i$ (excluding root variables) and (d) \texttt{NLL-Remainder:} Mean-NLL of all variables except root and intervention variables, see \Cref{fig:dissect_fig} for a graphical illustration.

Note that, causal models are not expected to yield comparable performance on  \texttt{NLL-Intervention} and \texttt{NLL-Root}, as causal models only use the parents of each variables to make its prediction. As this is an empty set for root variables and hard intervened variables, causal models estimate such variables from the marginal distribution of such variables. In contrast, monolithic models also rely on (potentially unstable) correlated predictors and anti-causal predictors and hence benefit from stronger performance as long none of these predictors were affected by an intervention.

\begin{figure}[t!]
    \centering
    \includegraphics[trim=0 45 0 0, clip, width=1.0\linewidth]{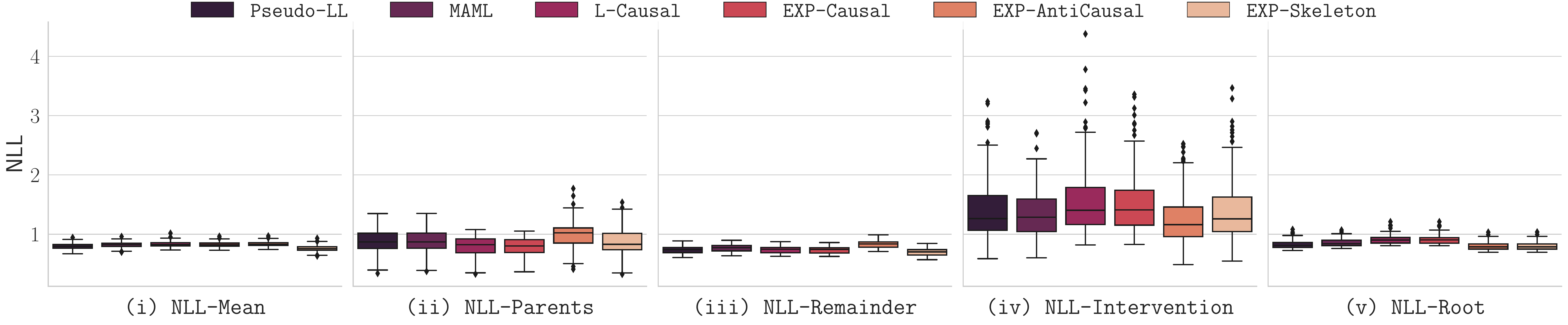}
    \includegraphics[trim=0 0 0 22, clip, width=1.0\linewidth]{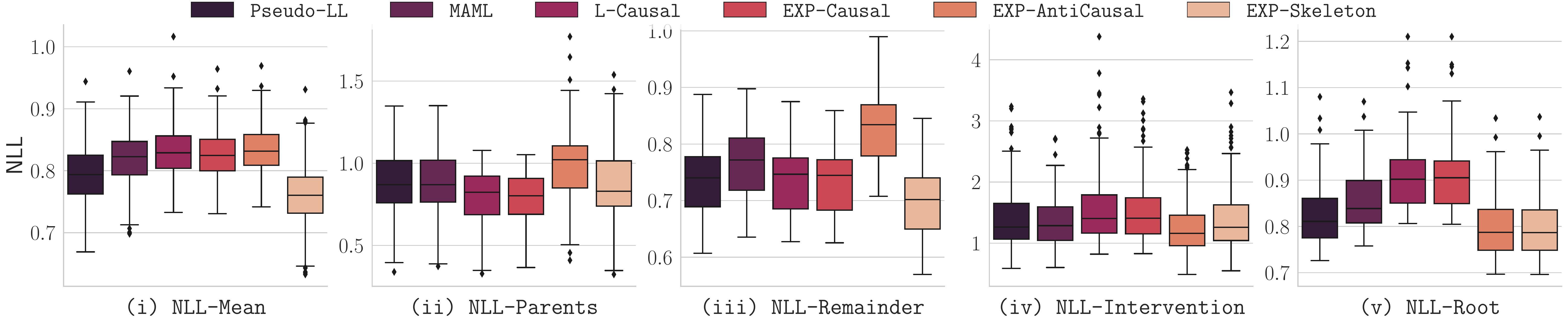}
    \caption{\textbf{NLL Dissection (Graph-Type: \texttt{ER-1}, $N=20$, Nr. Training Samples: $1000$).} Reporting the sub-metrics on the same scale (top row) clearly shows that \texttt{NLL-Parents} and \texttt{NLL-Intervention} are yielding NLL scores on a different scale. Therefore, we zoom in and show all sub-metrics on their own scale (bottom row). While all models achieve comparable results on most metrics, we observe that \textit{non-causal models} can catastrophically fail to predict the parent variables of an intervened variables (i.e. \texttt{NLL-Parents}). In contrast, \textit{causal models} maintain their performance and outperform all models on the \texttt{NLL-Parents} metric.  In general, we observe that \textit{causal models} yield robust performance across all sub-metrics. Furthermore, we observe advantages of \textit{non-causal models} over \textit{causal models} on the \texttt{NLL-Intervention} and \texttt{NLL-Root} metrics which is in line with our expectation as \textit{non-causal models} make use of non-causal predictors.}
    \label{fig:results_zeroshot_nllparents}
\end{figure}

\textbf{Summary.} Building upon our introduced sub-metrics, we find that important failure and robustness insights are hidden in a general evaluation scores such as \texttt{NLL-Mean} (see \Cref{fig:results_zeroshot_nllparents}). We observe different performance trends across all sub-metrics. While \textit{causal models} maintain a robust performance as they only rely on the inferred causal predictors, \textit{monolithic models} and \textit{structured models} that rely on anti-causal predictors (i.e. \texttt{EXP-AntiCausal} and \texttt{EXP-Skeleton}) show a significant deterioration in performance with large standard deviations on \texttt{NLL-Parents} as certain predictors got unstable due to the present distribution shift (i.e. modelled by a single-target intervention). We further observe stronger effects on sparse graphs where less stable predictors are available (see $\S$\ref{appendix:dissection_all}). Overall, this experiment confirms that all models expect \textit{causal models} show bigger risk for catastrophical failures by relying on unstable predictors.

\begin{figure}[t]
    \centering
    \includegraphics[width=0.49\linewidth]{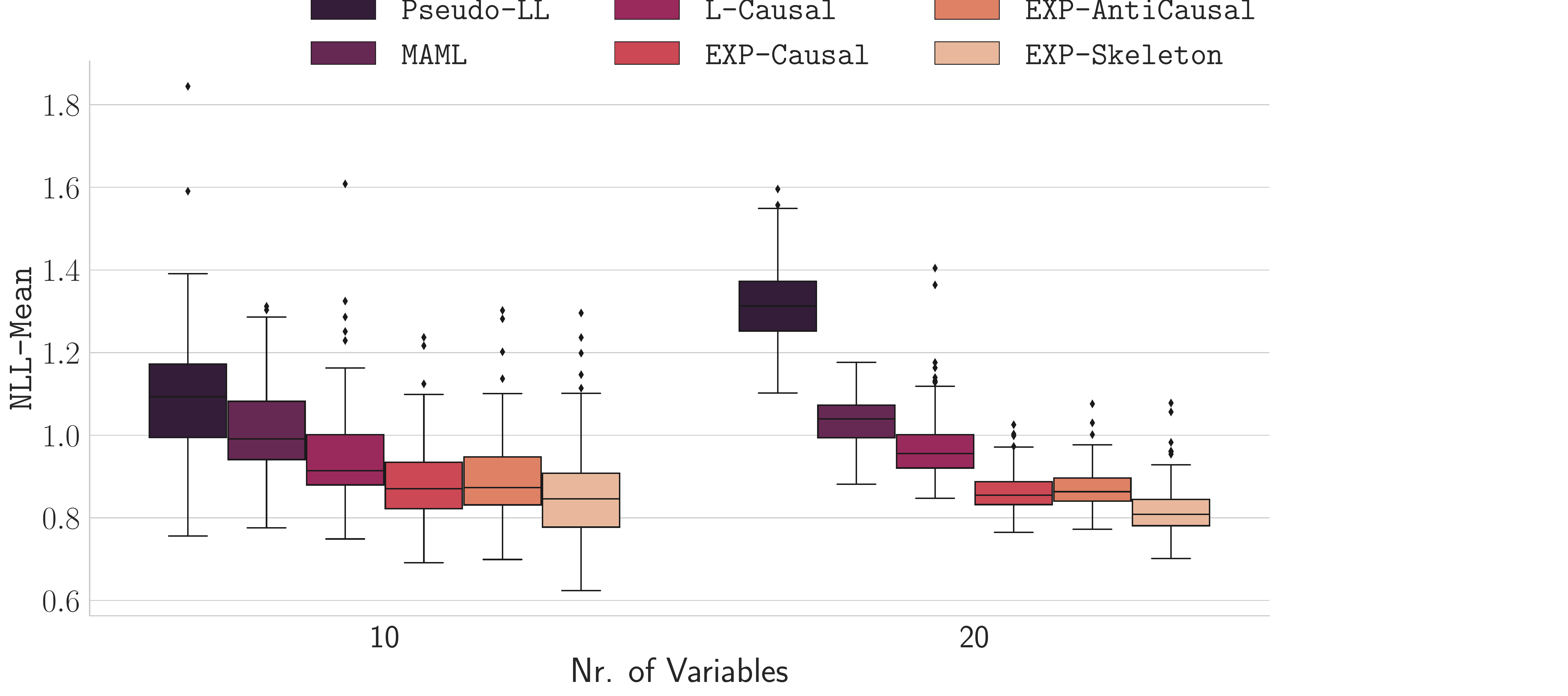}
    \hfill
    \includegraphics[width=0.49\linewidth]{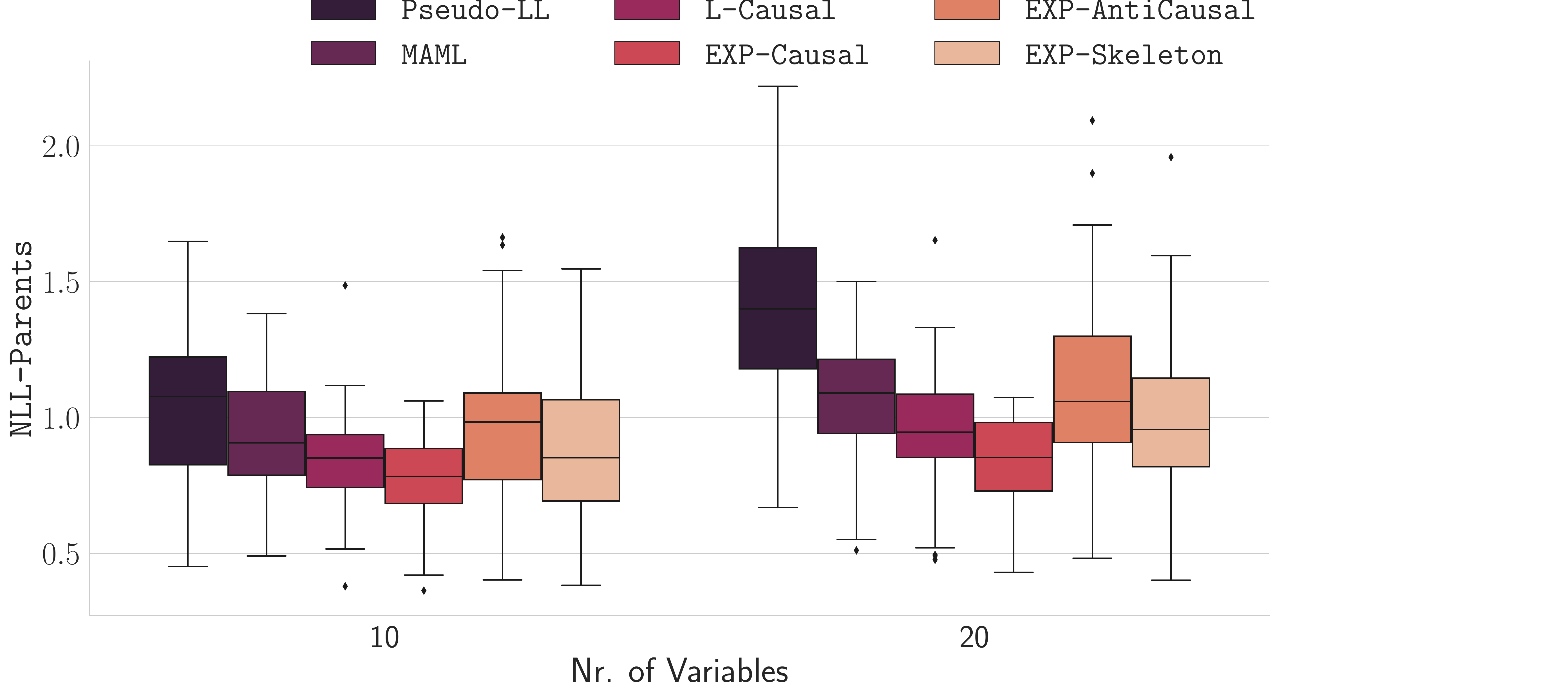}
    \caption{\textbf{Metric across increasing number of variables (fixed setting \texttt{ER-1} with 100 training samples).} We observe across all our experiments that the general generalization performance gaps with respect to \texttt{NLL-Mean} metric between \textit{monolithic models} and \textit{structured models} increases with increasing number of variables. On the same time, we observe a similar behavior on \texttt{NLL-Parents} where \textit{causal models} maintain robust generalization performance while all \textit{non-causal models} show big risks for catastrophic failure.}
    \label{fig:results_zeroshot_overNumberOfVariables}
    \vspace{-1\baselineskip}
\end{figure}

\vspace{-3mm}
\section{Analysis of Adaptation Performance}
\label{sec:adaptation}
In the previous section, we analyze the performance of different models in the zero-shot adaption setting; next, we analyze the performance of different models under the few-shot adaptation setting. The experiments in this section are designed to answer the following questions: (a). How well do different models adapt under different settings? (b) How the parameter space of different models are impacted by adaptation? (c) Can we leverage the the insights we learned from the previous experiment to improve adaptation speed for different models?

\textbf{Implementation details.} The model architecture and training setup is the same as in \Cref{sec:generalization}. For adaptation, all models are finetuned using stochastic gradient descent (SGD) with a step size of $0.1$.
\subsection{Adaptation performance}
We evaluate the adaption performance by evaluating how fast (speed of adaptation) and how well (how much overfitting) different models adapt to changes in distribution. We use two different adaptation techniques: (i) \textit{unconstrained adaptation} and (ii) \textit{sparse adaptation} (explained in \Cref{sec:adaptation_techniques}). Throughout this analysis, we pay particular attention to the \texttt{NLL-Parents} metrics and see how fast the affected models recover.

\begin{figure}[t]
    \centering
    \includegraphics[width=1.0\linewidth]{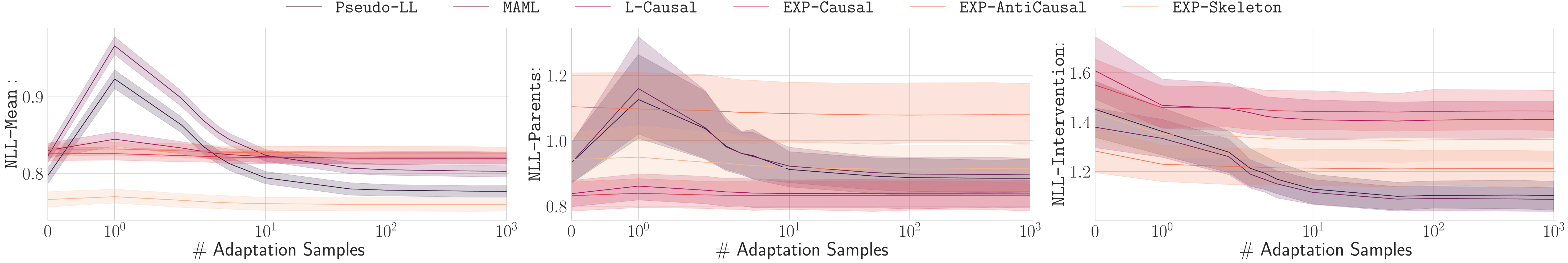}    
    \caption{\textbf{Speed of Adaptation in terms of different metrics.} \textit{Structured models} adapt considerably faster than \textit{monolithic models} across all settings and metrics. \textit{Monolithic models} show a sensitivity to overfitting if only low amounts of adaptation samples are available. We observe slightly faster adaptation for \textit{causal model} than for the other \textit{structured models}.}
    \label{fig:analysis_adaptation}
    \vspace{-1.2\baselineskip}
\end{figure}

\textbf{Speed of adaptation.}  
We fix the training data size of all models to be $10^3$ samples, as all models have converged on the generalization performance by then (\cref{fig:results_zeroshot_trainingData}).  We analyze the speed of adaption of different models by evaluating their adaption performance when finetuned using different amounts of data. Results shown in \cref{fig:analysis_adaptation}.

We observe that the \textit{structured models} adapt considerably faster than the \textit{monolithic models} across all settings. We observe slightly faster adaptation for \textit{causal model} than for the other \textit{structured models}. Within the \textit{monolithic models}, we observe that models that are trained with an adaptation objective (i.e. with MAML \citep{finn2017model}) adapt faster with respect to the intervened module than models trained on a naive pseudo-likelihood objective.

\textbf{Risk of overfitting.} Ideally, one aims to adapt to a transfer distribution using a less number of adaptation samples. In such a setting, a single update (i.e. gradient step with respect to the samples) using the available samples may be not sufficient to exploit all available information. Hence, it would be desirable to perform multiple updates on a small number of adaptation samples to extract all relevant information but without overfitting to the adaptation samples. To this end, we investigate the risk of overfitting when performing one or multiple updates on a fixed amount of data.

Across all our experiments, \textit{monolithic models} show strong overfitting effect when the number of adaptation samples are less, even on a single update step (see \cref{fig:analysis_adaptation}). In contrast, \textit{structured models} show reduced overfitting effects over multiple gradient steps, especially \textit{causal models}. For less number of adaptation samples, the speed of adaptation of \textit{causal models} can be further improved by employing a sparse adaptation objective. Overall, the adaptation landscape of the \textit{causal models} is significantly  \textit{different} from all other models, and hence allows to continuously improve the adaptation performance over multiple update steps.

\textbf{How does adaptation affect the parameter space $\theta$?} Based on the results from the previous analyses, we aim to further investigate the adaptation performance of models. We compare the effect on the parameters space $\theta$ between different models by employing the unconstrained adaptation objective, the most general adaptation objective.

\begin{figure}[t]
    \centering
     \includegraphics[width=1.0\textwidth]{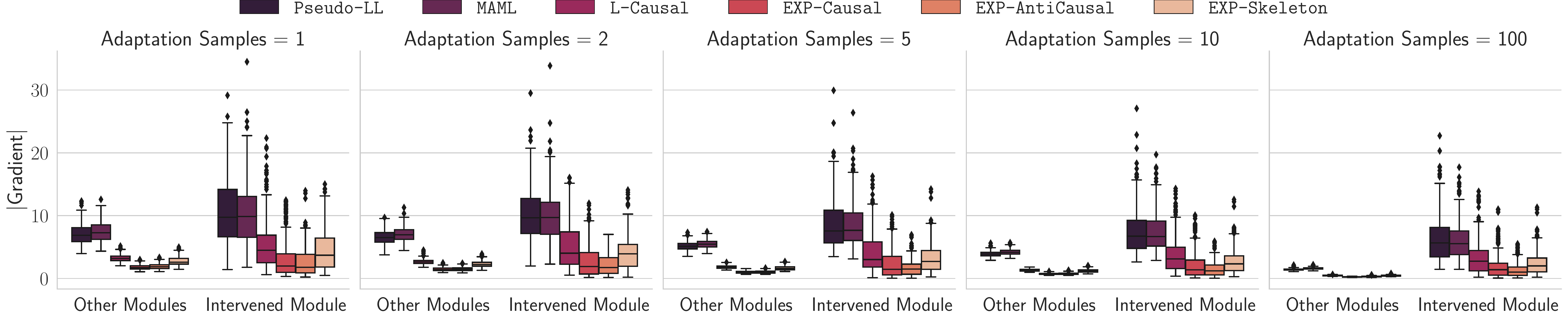}
    \caption{\textbf{Parameter Space Analysis.} While adapting to a shift in distribution, \textit{monolithic models} update most modules that were not affected by the intervention quite heavily compared to \textit{structured models}. \textit{Causal models} show remarkable adaptation behaviour in parameter space with localized updates on intervened modules.}
    \label{fig:analysis_parameterSpace}
    \vspace{-1.2\baselineskip}
\end{figure}

While \textit{monolithic models} adapt the parameters of many modules (i.e. independent  MLPs) heavily, measured with respect to the gradient magnitude, the adaptation of \textit{structured models} results in smaller updates, especially on the non-intervened modules (see \Cref{fig:analysis_parameterSpace}). \textit{Causal models} show remarkable adaptation behaviour in parameter space with \textit{localized updates on certain modules}, and significantly reduced gradient magnitudes. If a \textit{causal model} is trained on enough training samples and has access to enough adaptation adaptation samples, the unconstrained adaptation without knowledge about the intervention target yields nearly the same update as if we enforce the sparse update on the known intervened module (see \S \ref{appendix:parameter_space_analysis}).

\vspace{-2mm}

\subsection{How to adapt causal models efficiently in all settings?}
\vspace{-2mm}
\begin{wrapfigure}{r}{0.35\textwidth}
    \centering
     \includegraphics[width=0.34\textwidth]{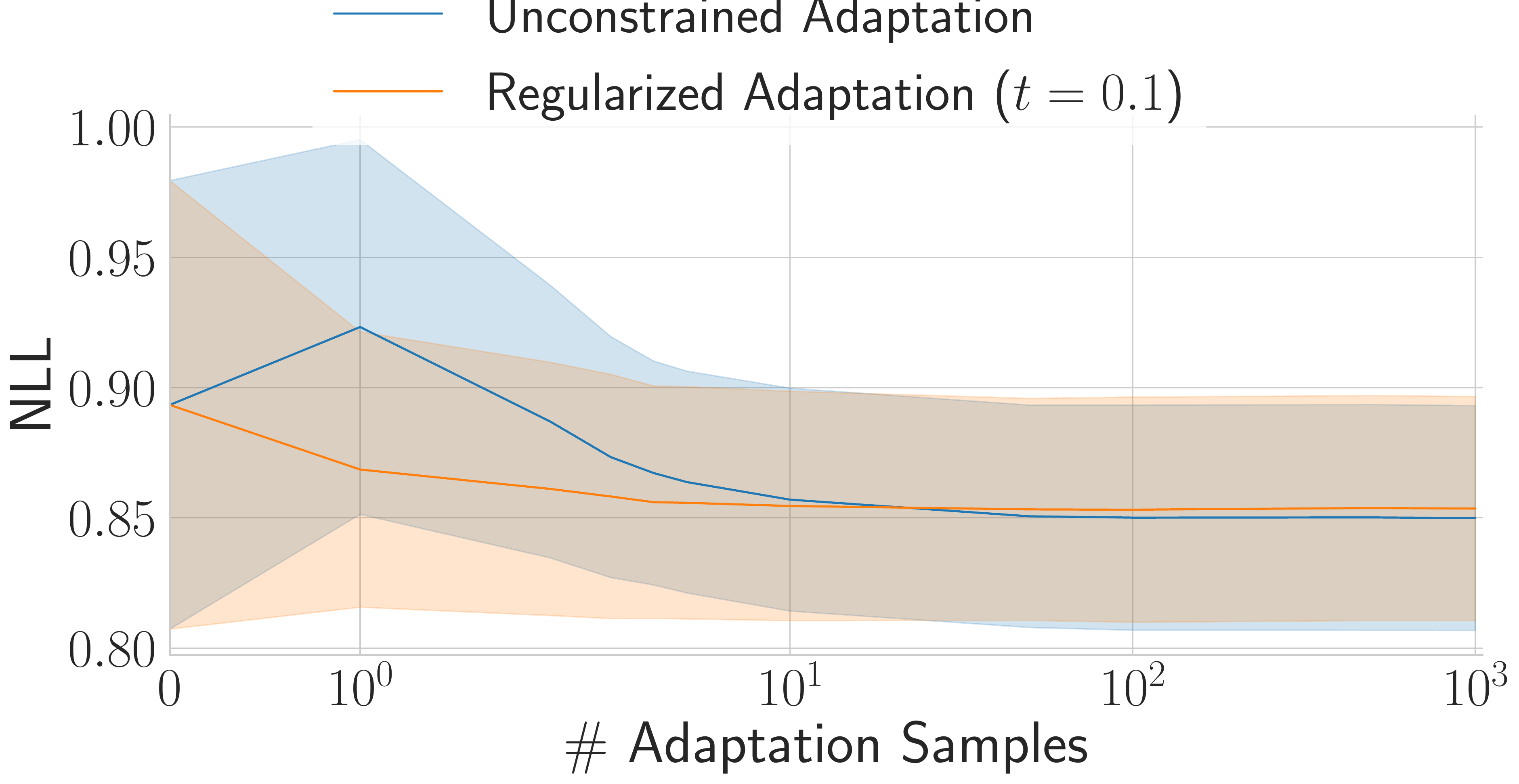}
    \caption{\textbf{Efficient Adaptation of a causal model.} Regularized Adaptation leads to efficient adaptation with low amounts of samples and  improves the speed of adaptation compared to unconstrained adaptation. \\}
    \label{fig:regularizedAdapt}
\end{wrapfigure}

So far, we have observed the effects of an unconstrained adaptation objective where all modules can be updated, and the sparse adaptation objective where only a specific module is updated (i.e. module is either estimated or known). However, it would be desirable to update the modules in a more efficient manner using less amount of adaptation samples without overfitting. To this end, we employ our proposed regularized adaptation objective and investigate the effect on the speed of adaptation. 

Across our  experiments, we observed how the proposed adaptation objective further improves the speed of adaptation if only less amount of adaptation samples are available (as shown in \Cref{fig:regularizedAdapt}). It especially improves adaptation if the pretrained model is only trained on few amounts of data. Further, it improves the statistical efficiency compared to the sparse adaptation objective, as available samples are used to update necessary module, if  necessary.

\vspace{-1mm}
\section{Conclusion}
In this work, we systematically analyzed the generalization and adaptation performance of different models ranging from \textit{monolithic models} that have no inbuilt structure to \textit{structured models} that are either provided with structural expert knowledge upfront or learn structure from data. Our experiments show that the \textit{causal models} significantly outperform \textit{non-causal} models in low-data regimes and offer robust generalization across all settings. In a further analysis, we evaluated few-shot adaptation in various settings and show that \textit{causal models} offer fast and robust adaptation with only less number of adaptation samples. Based on these results, we analyze how the adaptation performance relates to changes in the parameter space and then proposed  a new adaptation objective that dynamically modulates the degree of adaptation and hence allows more sample efficient adaptation. In this work, we considered relatively low-dimensional settings where causal variables are explicitly given. Translating our systematic evaluation and score dissection analysis to high-dimensional evaluation setups such as \citep{ke2021systematic} would be an interesting direction for future work.

\textbf{Limitations.} In the present work, we have only experimented with one specific class of a neural causal discovery framework \citep{ke2019learning, lippe2021efficient} to learn causal structure from data. Hence, the performance of the learned causal model may vary with other classes of neural causal discovery frameworks. However, we have introduced a causal model where the true causal structure is presented upfront and hence represents an performance upper-bound given a certain amount of training data. Furthermore, we have only conducted experiments on datasets where the causal variables are explicitly observed and the underlying causal graph is acyclic.

\begin{ack}
The authors would like to acknowledge the support of the following agencies for research funding
and computing support: Compute Canada, the Canada Research Chairs, CIFAR. Yoshua Bengio is a CIFAR Senior Fellow and Stefan Bauer is a CIFAR Azrieli Global.
\end{ack}

\medskip
\bibliographystyle{plainnat}
\bibliography{main}

\begin{thebibliography}{49}
\providecommand{\natexlab}[1]{#1}
\providecommand{\url}[1]{\texttt{#1}}
\expandafter\ifx\csname urlstyle\endcsname\relax
  \providecommand{\doi}[1]{doi: #1}\else
  \providecommand{\doi}{doi: \begingroup \urlstyle{rm}\Url}\fi

\bibitem[Alet et~al.(2018)Alet, Lozano-P{\'e}rez, and
  Kaelbling]{alet2018modular}
Ferran Alet, Tom{\'a}s Lozano-P{\'e}rez, and Leslie~P Kaelbling.
\newblock Modular meta-learning.
\newblock In \emph{Conference on Robot Learning}, pages 856--868. PMLR, 2018.

\bibitem[Annadani et~al.(2021)Annadani, Rothfuss, Lacoste, Scherrer, Goyal,
  Bengio, and Bauer]{annadani2021variational}
Yashas Annadani, Jonas Rothfuss, Alexandre Lacoste, Nino Scherrer, Anirudh
  Goyal, Yoshua Bengio, and Stefan Bauer.
\newblock Variational causal networks: Approximate bayesian inference over
  causal structures.
\newblock \emph{arXiv preprint arXiv:2106.07635}, 2021.

\bibitem[Bahdanau et~al.(2018)Bahdanau, Murty, Noukhovitch, Nguyen, de~Vries,
  and Courville]{bahdanau2018systematic}
Dzmitry Bahdanau, Shikhar Murty, Michael Noukhovitch, Thien~Huu Nguyen, Harm
  de~Vries, and Aaron Courville.
\newblock Systematic generalization: what is required and can it be learned?
\newblock \emph{arXiv preprint arXiv:1811.12889}, 2018.

\bibitem[Bareinboim and Pearl(2016)]{bareinboim2016causal}
Elias Bareinboim and Judea Pearl.
\newblock Causal inference and the data-fusion problem.
\newblock \emph{Proceedings of the National Academy of Sciences}, 113\penalty0
  (27):\penalty0 7345--7352, 2016.

\bibitem[Bengio et~al.(2019)Bengio, Deleu, Rahaman, Ke, Lachapelle, Bilaniuk,
  Goyal, and Pal]{bengio2019meta}
Yoshua Bengio, Tristan Deleu, Nasim Rahaman, Rosemary Ke, S{\'e}bastien
  Lachapelle, Olexa Bilaniuk, Anirudh Goyal, and Christopher Pal.
\newblock A meta-transfer objective for learning to disentangle causal
  mechanisms.
\newblock \emph{arXiv preprint arXiv:1901.10912}, 2019.

\bibitem[Brouillard et~al.(2020)Brouillard, Lachapelle, Lacoste,
  Lacoste-Julien, and Drouin]{brouillard2020differentiable}
Philippe Brouillard, S\'{e}bastien Lachapelle, Alexandre Lacoste, Simon
  Lacoste-Julien, and Alexandre Drouin.
\newblock Differentiable causal discovery from interventional data.
\newblock In H.~Larochelle, M.~Ranzato, R.~Hadsell, M.~F. Balcan, and H.~Lin,
  editors, \emph{Advances in Neural Information Processing Systems}, volume~33,
  pages 21865--21877. Curran Associates, Inc., 2020.

\bibitem[Chen et~al.(2020)Chen, Friesen, Behbahani, Doucet, Budden, Hoffman,
  and de~Freitas]{chen2020modular}
Yutian Chen, Abram~L Friesen, Feryal Behbahani, Arnaud Doucet, David Budden,
  Matthew Hoffman, and Nando de~Freitas.
\newblock Modular meta-learning with shrinkage.
\newblock \emph{Advances in Neural Information Processing Systems},
  33:\penalty0 2858--2869, 2020.

\bibitem[Cobbe et~al.(2019)Cobbe, Klimov, Hesse, Kim, and
  Schulman]{cobbe2019quantifying}
Karl Cobbe, Oleg Klimov, Chris Hesse, Taehoon Kim, and John Schulman.
\newblock Quantifying generalization in reinforcement learning.
\newblock In \emph{International Conference on Machine Learning}, pages
  1282--1289. PMLR, 2019.

\bibitem[Cundy et~al.(2021)Cundy, Grover, and Ermon]{cundy2021bcd}
Chris Cundy, Aditya Grover, and Stefano Ermon.
\newblock Bcd nets: Scalable variational approaches for bayesian causal
  discovery.
\newblock \emph{Advances in Neural Information Processing Systems}, 34, 2021.

\bibitem[Dasgupta et~al.(2019)Dasgupta, Wang, Chiappa, Mitrovic, Ortega,
  Raposo, Hughes, Battaglia, Botvinick, and Kurth-Nelson]{dasgupta2019causal}
Ishita Dasgupta, Jane Wang, Silvia Chiappa, Jovana Mitrovic, Pedro Ortega,
  David Raposo, Edward Hughes, Peter Battaglia, Matthew Botvinick, and Zeb
  Kurth-Nelson.
\newblock Causal {R}easoning from {M}eta-reinforcement {L}earning.
\newblock \emph{arXiv preprint arXiv:1901.08162}, 2019.

\bibitem[De~Haan et~al.(2019)De~Haan, Jayaraman, and Levine]{de2019causal}
Pim De~Haan, Dinesh Jayaraman, and Sergey Levine.
\newblock Causal confusion in imitation learning.
\newblock \emph{Advances in Neural Information Processing Systems}, 32, 2019.

\bibitem[Deleu et~al.(2019)Deleu, W\"urfl, Samiei, Cohen, and
  Bengio]{deleu2019torchmeta}
Tristan Deleu, Tobias W\"urfl, Mandana Samiei, Joseph~Paul Cohen, and Yoshua
  Bengio.
\newblock {Torchmeta: A Meta-Learning library for PyTorch}, 2019.
\newblock URL \url{https://arxiv.org/abs/1909.06576}.
\newblock Available at: https://github.com/tristandeleu/pytorch-meta.

\bibitem[Djolonga et~al.(2021)Djolonga, Yung, Tschannen, Romijnders, Beyer,
  Kolesnikov, Puigcerver, Minderer, D'Amour, Moldovan,
  et~al.]{djolonga2021robustness}
Josip Djolonga, Jessica Yung, Michael Tschannen, Rob Romijnders, Lucas Beyer,
  Alexander Kolesnikov, Joan Puigcerver, Matthias Minderer, Alexander D'Amour,
  Dan Moldovan, et~al.
\newblock On robustness and transferability of convolutional neural networks.
\newblock In \emph{Proceedings of the IEEE/CVF Conference on Computer Vision
  and Pattern Recognition}, pages 16458--16468, 2021.

\bibitem[Finn et~al.(2017)Finn, Abbeel, and Levine]{finn2017model}
Chelsea Finn, Pieter Abbeel, and Sergey Levine.
\newblock Model-agnostic meta-learning for fast adaptation of deep networks.
\newblock In \emph{International conference on machine learning}, pages
  1126--1135. PMLR, 2017.

\bibitem[Geffner et~al.(2022)Geffner, Antoran, Foster, Gong, Ma, Kiciman,
  Sharma, Lamb, Kukla, Pawlowski, et~al.]{geffner2022deep}
Tomas Geffner, Javier Antoran, Adam Foster, Wenbo Gong, Chao Ma, Emre Kiciman,
  Amit Sharma, Angus Lamb, Martin Kukla, Nick Pawlowski, et~al.
\newblock Deep end-to-end causal inference.
\newblock \emph{arXiv preprint arXiv:2202.02195}, 2022.

\bibitem[Goyal et~al.(2019)Goyal, Lamb, Hoffmann, Sodhani, Levine, Bengio, and
  Sch{\"o}lkopf]{goyal2019recurrent}
Anirudh Goyal, Alex Lamb, Jordan Hoffmann, Shagun Sodhani, Sergey Levine,
  Yoshua Bengio, and Bernhard Sch{\"o}lkopf.
\newblock Recurrent independent mechanisms.
\newblock \emph{arXiv preprint arXiv:1909.10893}, 2019.

\bibitem[Ke et~al.(2019)Ke, Bilaniuk, Goyal, Bauer, Larochelle, Sch{\"o}lkopf,
  Mozer, Pal, and Bengio]{ke2019learning}
Nan~Rosemary Ke, Olexa Bilaniuk, Anirudh Goyal, Stefan Bauer, Hugo Larochelle,
  Bernhard Sch{\"o}lkopf, Michael~C Mozer, Chris Pal, and Yoshua Bengio.
\newblock Learning neural causal models from unknown interventions.
\newblock \emph{arXiv preprint arXiv:1910.01075}, 2019.

\bibitem[Ke et~al.(2021)Ke, Didolkar, Mittal, Goyal, Lajoie, Bauer, Rezende,
  Bengio, Mozer, and Pal]{ke2021systematic}
Nan~Rosemary Ke, Aniket Didolkar, Sarthak Mittal, Anirudh Goyal, Guillaume
  Lajoie, Stefan Bauer, Danilo Rezende, Yoshua Bengio, Michael Mozer, and
  Christopher Pal.
\newblock Systematic evaluation of causal discovery in visual model based
  reinforcement learning.
\newblock \emph{arXiv preprint arXiv:2107.00848}, 2021.

\bibitem[Ke et~al.(2022)Ke, Chiappa, Wang, Bornschein, Weber, Goyal, Botvinic,
  Mozer, and Rezende]{ke2022learning}
Nan~Rosemary Ke, Silvia Chiappa, Jane Wang, Jorg Bornschein, Theophane Weber,
  Anirudh Goyal, Matthew Botvinic, Michael Mozer, and Danilo~Jimenez Rezende.
\newblock Learning to induce causal structure.
\newblock \emph{arXiv preprint arXiv:2204.04875}, 2022.

\bibitem[Kingma and Ba(2014)]{kingma2014adam}
Diederik~P Kingma and Jimmy Ba.
\newblock Adam: A method for stochastic optimization.
\newblock \emph{arXiv preprint arXiv:1412.6980}, 2014.

\bibitem[Koh et~al.(2021)Koh, Sagawa, Marklund, Xie, Zhang, Balsubramani, Hu,
  Yasunaga, Phillips, Gao, et~al.]{koh2021wilds}
Pang~Wei Koh, Shiori Sagawa, Henrik Marklund, Sang~Michael Xie, Marvin Zhang,
  Akshay Balsubramani, Weihua Hu, Michihiro Yasunaga, Richard~Lanas Phillips,
  Irena Gao, et~al.
\newblock Wilds: A benchmark of in-the-wild distribution shifts.
\newblock In \emph{International Conference on Machine Learning}, pages
  5637--5664. PMLR, 2021.

\bibitem[Kyono et~al.(2020)Kyono, Zhang, and van~der Schaar]{kyono2020castle}
Trent Kyono, Yao Zhang, and Mihaela van~der Schaar.
\newblock Castle: Regularization via auxiliary causal graph discovery.
\newblock \emph{Advances in Neural Information Processing Systems},
  33:\penalty0 1501--1512, 2020.

\bibitem[Lachapelle et~al.(2019)Lachapelle, Brouillard, Deleu, and
  Lacoste-Julien]{lachapelle2019gradient}
S{\'e}bastien Lachapelle, Philippe Brouillard, Tristan Deleu, and Simon
  Lacoste-Julien.
\newblock Gradient-based neural dag learning.
\newblock \emph{arXiv preprint arXiv:1906.02226}, 2019.

\bibitem[Lake et~al.(2017)Lake, Ullman, Tenenbaum, and
  Gershman]{lake2017building}
Brenden~M Lake, Tomer~D Ullman, Joshua~B Tenenbaum, and Samuel~J Gershman.
\newblock Building machines that learn and think like people.
\newblock \emph{Behavioral and brain sciences}, 40, 2017.

\bibitem[Le~Priol et~al.(2021)Le~Priol, Babanezhad, Bengio, and
  Lacoste-Julien]{le2021analysis}
R{\'e}mi Le~Priol, Reza Babanezhad, Yoshua Bengio, and Simon Lacoste-Julien.
\newblock An analysis of the adaptation speed of causal models.
\newblock In \emph{International Conference on Artificial Intelligence and
  Statistics}, pages 775--783. PMLR, 2021.

\bibitem[Lippe et~al.(2021)Lippe, Cohen, and Gavves]{lippe2021efficient}
Phillip Lippe, Taco Cohen, and Efstratios Gavves.
\newblock Efficient neural causal discovery without acyclicity constraints.
\newblock \emph{arXiv preprint arXiv:2107.10483}, 2021.

\bibitem[Lorch et~al.(2021)Lorch, Rothfuss, Sch{\"o}lkopf, and
  Krause]{lorch2021dibs}
Lars Lorch, Jonas Rothfuss, Bernhard Sch{\"o}lkopf, and Andreas Krause.
\newblock Dibs: Differentiable bayesian structure learning.
\newblock \emph{arXiv preprint arXiv:2105.11839}, 2021.

\bibitem[Madan et~al.(2021)Madan, Ke, Goyal, Sch{\"o}lkopf, and
  Bengio]{madan2021fast}
Kanika Madan, Nan~Rosemary Ke, Anirudh Goyal, Bernhard Sch{\"o}lkopf, and
  Yoshua Bengio.
\newblock Fast and slow learning of recurrent independent mechanisms.
\newblock \emph{arXiv preprint arXiv:2105.08710}, 2021.

\bibitem[Magliacane et~al.(2018)Magliacane, van Ommen, Claassen, Bongers,
  Versteeg, and Mooij]{magliacane2018domain}
Sara Magliacane, Thijs van Ommen, Tom Claassen, Stephan Bongers, Philip
  Versteeg, and Joris~M Mooij.
\newblock Domain adaptation by using causal inference to predict invariant
  conditional distributions.
\newblock In \emph{Advances in Neural Information Processing Systems}, pages
  10869--10879, 2018.

\bibitem[Nair et~al.(2019)Nair, Zhu, Savarese, and Fei-Fei]{nair2019causal}
Suraj Nair, Yuke Zhu, Silvio Savarese, and Li~Fei-Fei.
\newblock Causal induction from visual observations for goal directed tasks.
\newblock \emph{arXiv preprint arXiv:1910.01751}, 2019.

\bibitem[Nichol et~al.(2018{\natexlab{a}})Nichol, Achiam, and
  Schulman]{nichol2018first}
Alex Nichol, Joshua Achiam, and John Schulman.
\newblock On first-order meta-learning algorithms.
\newblock \emph{arXiv preprint arXiv:1803.02999}, 2018{\natexlab{a}}.

\bibitem[Nichol et~al.(2018{\natexlab{b}})Nichol, Pfau, Hesse, Klimov, and
  Schulman]{nichol2018gotta}
Alex Nichol, Vicki Pfau, Christopher Hesse, Oleg Klimov, and John Schulman.
\newblock Gotta learn fast: A new benchmark for generalization in rl.
\newblock \emph{arXiv preprint arXiv:1804.03720}, 2018{\natexlab{b}}.

\bibitem[Packer et~al.(2018)Packer, Gao, Kos, Kr{\"a}henb{\"u}hl, Koltun, and
  Song]{packer2018assessing}
Charles Packer, Katelyn Gao, Jernej Kos, Philipp Kr{\"a}henb{\"u}hl, Vladlen
  Koltun, and Dawn Song.
\newblock Assessing generalization in deep reinforcement learning.
\newblock \emph{arXiv preprint arXiv:1810.12282}, 2018.

\bibitem[Parascandolo et~al.(2018)Parascandolo, Kilbertus, Rojas-Carulla, and
  Sch{\"o}lkopf]{parascandolo2018learning}
Giambattista Parascandolo, Niki Kilbertus, Mateo Rojas-Carulla, and Bernhard
  Sch{\"o}lkopf.
\newblock Learning independent causal mechanisms.
\newblock In \emph{International Conference on Machine Learning}, pages
  4036--4044. PMLR, 2018.

\bibitem[Pearl(1995)]{pearl1995causal}
Judea Pearl.
\newblock Causal diagrams for empirical research.
\newblock \emph{Biometrika}, 82\penalty0 (4):\penalty0 669--688, 1995.

\bibitem[Peters et~al.(2016)Peters, B{\"u}hlmann, and
  Meinshausen]{peters2016causal}
Jonas Peters, Peter B{\"u}hlmann, and Nicolai Meinshausen.
\newblock Causal inference by using invariant prediction: identification and
  confidence intervals.
\newblock \emph{Journal of the Royal Statistical Society: Series B (Statistical
  Methodology)}, 78\penalty0 (5):\penalty0 947--1012, 2016.

\bibitem[Peters et~al.(2017)Peters, Janzing, and
  Sch{\"o}lkopf]{peters2017elements}
Jonas Peters, Dominik Janzing, and Bernhard Sch{\"o}lkopf.
\newblock \emph{Elements of causal inference: foundations and learning
  algorithms}.
\newblock The MIT Press, 2017.

\bibitem[Rezende et~al.(2020)Rezende, Danihelka, Papamakarios, Ke, Jiang,
  Weber, Gregor, Merzic, Viola, Wang, et~al.]{rezende2020causally}
Danilo~J Rezende, Ivo Danihelka, George Papamakarios, Nan~Rosemary Ke, Ray
  Jiang, Theophane Weber, Karol Gregor, Hamza Merzic, Fabio Viola, Jane Wang,
  et~al.
\newblock Causally correct partial models for reinforcement learning.
\newblock \emph{arXiv preprint arXiv:2002.02836}, 2020.

\bibitem[Rojas-Carulla et~al.(2018)Rojas-Carulla, Sch{\"o}lkopf, Turner, and
  Peters]{rojas2018invariant}
Mateo Rojas-Carulla, Bernhard Sch{\"o}lkopf, Richard Turner, and Jonas Peters.
\newblock Invariant models for causal transfer learning.
\newblock \emph{The Journal of Machine Learning Research}, 19\penalty0
  (1):\penalty0 1309--1342, 2018.

\bibitem[Rosenfeld et~al.(2018)Rosenfeld, Zemel, and
  Tsotsos]{rosenfeld2018elephant}
Amir Rosenfeld, Richard Zemel, and John~K Tsotsos.
\newblock The elephant in the room.
\newblock \emph{arXiv preprint arXiv:1808.03305}, 2018.

\bibitem[Scherrer et~al.(2021)Scherrer, Bilaniuk, Annadani, Goyal, Schwab,
  Sch{\"o}lkopf, Mozer, Bengio, Bauer, and Ke]{scherrer2021learning}
Nino Scherrer, Olexa Bilaniuk, Yashas Annadani, Anirudh Goyal, Patrick Schwab,
  Bernhard Sch{\"o}lkopf, Michael~C Mozer, Yoshua Bengio, Stefan Bauer, and
  Nan~Rosemary Ke.
\newblock Learning neural causal models with active interventions.
\newblock \emph{arXiv preprint arXiv:2109.02429}, 2021.

\bibitem[Sch{\"o}lkopf et~al.(2021)Sch{\"o}lkopf, Locatello, Bauer, Ke,
  Kalchbrenner, Goyal, and Bengio]{scholkopf2021toward}
Bernhard Sch{\"o}lkopf, Francesco Locatello, Stefan Bauer, Nan~Rosemary Ke, Nal
  Kalchbrenner, Anirudh Goyal, and Yoshua Bengio.
\newblock Toward causal representation learning.
\newblock \emph{Proceedings of the IEEE}, 109\penalty0 (5):\penalty0 612--634,
  2021.

\bibitem[Taori et~al.(2020)Taori, Dave, Shankar, Carlini, Recht, and
  Schmidt]{taori2020measuring}
Rohan Taori, Achal Dave, Vaishaal Shankar, Nicholas Carlini, Benjamin Recht,
  and Ludwig Schmidt.
\newblock Measuring robustness to natural distribution shifts in image
  classification.
\newblock \emph{Advances in Neural Information Processing Systems},
  33:\penalty0 18583--18599, 2020.

\bibitem[Tigas et~al.(2022)Tigas, Annadani, Jesson, Sch{\"o}lkopf, Gal, and
  Bauer]{tigas2022interventions}
Panagiotis Tigas, Yashas Annadani, Andrew Jesson, Bernhard Sch{\"o}lkopf, Yarin
  Gal, and Stefan Bauer.
\newblock Interventions, where and how? experimental design for causal models
  at scale.
\newblock \emph{arXiv preprint arXiv:2203.02016}, 2022.

\bibitem[Wang et~al.(2021)Wang, Lan, Liu, Ouyang, Zeng, and
  Qin]{wang2021generalizing}
Jindong Wang, Cuiling Lan, Chang Liu, Yidong Ouyang, Wenjun Zeng, and Tao Qin.
\newblock Generalizing to unseen domains: A survey on domain generalization.
\newblock \emph{arXiv preprint arXiv:2103.03097}, 2021.

\bibitem[Yu et~al.(2019)Yu, Chen, Gao, and Yu]{yu2019dag}
Yue Yu, Jie Chen, Tian Gao, and Mo~Yu.
\newblock Dag-gnn: Dag structure learning with graph neural networks.
\newblock In \emph{International Conference on Machine Learning}, pages
  7154--7163. PMLR, 2019.

\bibitem[Zhang et~al.(2020)Zhang, Zhang, and Li]{zhang2020causal}
Cheng Zhang, Kun Zhang, and Yingzhen Li.
\newblock A causal view on robustness of neural networks.
\newblock \emph{Advances in Neural Information Processing Systems},
  33:\penalty0 289--301, 2020.

\bibitem[Zhang et~al.(2013)Zhang, Sch{\"o}lkopf, Muandet, and
  Wang]{zhang2013domain}
Kun Zhang, Bernhard Sch{\"o}lkopf, Krikamol Muandet, and Zhikun Wang.
\newblock Domain adaptation under target and conditional shift.
\newblock In \emph{International Conference on Machine Learning}, pages
  819--827. PMLR, 2013.

\bibitem[Zheng et~al.(2018)Zheng, Aragam, Ravikumar, and Xing]{zheng2018dags}
Xun Zheng, Bryon Aragam, Pradeep~K Ravikumar, and Eric~P Xing.
\newblock {DAGs} with {NO TEARS}: Continuous optimization for structure
  learning.
\newblock In \emph{Advances in Neural Information Processing Systems},
  volume~31, pages 9472--9483, 2018.

\end{thebibliography}


\if{0}
    \clearpage
    \section*{Checklist}


    \begin{enumerate}

    \item For all authors...
    \begin{enumerate}
      \item Do the main claims made in the abstract and introduction accurately reflect the paper's contributions and scope?
        \answerYes{}
      \item Did you describe the limitations of your work?
        \answerYes{See \Cref{appendix:limitations}}
      \item Did you discuss any potential negative societal impacts of your work?
        \answerNA{Besides the general impact of machine learning methods we do not foresee and identify any particular and immediate negative societal impact of our work. We do not conduct any experiments that could be in itself harmful for society but understand that the applicability of this method is very broad.}
      \item Have you read the ethics review guidelines and ensured that your paper conforms to them?
        \answerYes{}
    \end{enumerate}

    \item If you are including theoretical results...
    \begin{enumerate}
      \item Did you state the full set of assumptions of all theoretical results?
        \answerNA
            \item Did you include complete proofs of all theoretical results?
        \answerNA
    \end{enumerate}

    \item If you ran experiments...
    \begin{enumerate}
      \item Did you include the code, data, and instructions needed to reproduce the main experimental results (either in the supplemental material or as a URL)?
        \answerYes{We will include the code in supplementary material.}
      \item Did you specify all the training details (e.g., data splits, hyperparameters, how they were chosen)?
        \answerYes{We specified the most important training details  and additional implementation details in the appendix.}
            \item Did you report error bars (e.g., with respect to the random seed after running experiments multiple times)?
        \answerYes{We train every model with 10 random seeds each.}
            \item Did you include the total amount of compute and the type of resources used (e.g., type of GPUs, internal cluster, or cloud provider)?
        \answerYes{We provide an approximate estimate of the overall required compute in the appendix.}
    \end{enumerate}

    \item If you are using existing assets (e.g., code, data, models) or curating/releasing new assets...
    \begin{enumerate}
      \item If your work uses existing assets, did you cite the creators?
        \answerYes{We cite the creators of all datasets and pre-trained models being used.}
      \item Did you mention the license of the assets?
        \answerYes{As stated in the appendix, all software and assets we use are open source and under MIT, Apache or Creative Commons Licenses.}
      \item Did you include any new assets either in the supplemental material or as a URL?
        \answerNA{}
      \item Did you discuss whether and how consent was obtained from people whose data you're using/curating?
        \answerNA{We use benchmark datasets that are widely used for ML research and cite them accordingly.}
      \item Did you discuss whether the data you are using/curating contains personally identifiable information or offensive content?
        \answerNA{We do not use any data that contains personally identifiable information.}
    \end{enumerate}

    \item If you used crowdsourcing or conducted research with human subjects...
    \begin{enumerate}
      \item Did you include the full text of instructions given to participants and screenshots, if applicable?
        \answerNA{}
      \item Did you describe any potential participant risks, with links to Institutional Review Board (IRB) approvals, if applicable?
        \answerNA{}
      \item Did you include the estimated hourly wage paid to participants and the total amount spent on participant compensation?
        \answerNA{}
    \end{enumerate}

    \end{enumerate}
\fi

\clearpage
\appendix
\section{Appendix}

\vspace{5mm}
\vskip3pt\hrule\vskip5pt
\localtableofcontents
\vskip3pt\hrule\vskip5pt
\clearpage


\clearpage

\subsection{Use of Existing Assets}
In the present work, we made partially use of following assets:
\begin{itemize}
    \item SDI \citep{ke2019learning}: \href{https://github.com/nke001/causal_learning\_unknown\_interventions}{https://github.com/nke001/causal\_learning\_unknown\_interventions}
    \item ENCO \citep{lippe2021efficient}: \href{https://github.com/phlippe/ENCO}{https://github.com/phlippe/ENCO}
    \item TorchMeta \citep{deleu2019torchmeta}: \href{https://github.com/tristandeleu/pytorch-meta}{https://github.com/tristandeleu/pytorch-meta}
\end{itemize}

\subsection{Technical Details}
\subsubsection{Detailed Training Objective: Learning Causal Structure from Data \texttt{L-Causal}}
\label{appendix:implementation_modelArchitecture}

We use a  causal discovery framework to learn a structural causal model (SCM) from data. To this end, we follow the setup of \citet{lippe2021efficient} and introduce an additional set of parameters $\gamma = (u, v)$ with $u \in \mathbb{R}^{N \times N}$ and $v \in \mathbb{R}^{N \times N}$ which define a continuous relaxation of an adjacency matrix $\gamma = \sigma(u) \cdot \sigma(v)$. Such a soft-adjacency matrix can be conveniently used to sample input masks $M$. In order to train the parameters $\theta$ of the MLPs  and the learnable input mask $\gamma$, the framework relies on a optimization formulation using two alternating phases of optimization \citep{ke2019learning, lippe2021efficient, scherrer2021learning}. These are performed until convergence in an iterative manner. Under freezed mask parameters $\gamma$, we train during phase 1 (called "Distribution Fitting") the parameters $\theta_i$ of each MLP on observational data $\mathcal{D}_{obs}$ using a similar maximum likelihood objective as in \Cref{eq:MLE}: 
    \begin{equation}
        \small
        \theta_i^{*} = \argmin_{\theta_i} \mathop{\mathbb{E}}_{X\sim\mathcal{D}_{obs}} \mathop{\mathbb{E}}_{M \sim p(M;u,v)}[-log(f(X, M_i;\theta_i))]
    \end{equation}
    where we sample a set of input masks $M$ from $p(M;u,v)$ instead of relying on a fixed mask $M$, where $M_{ij} \sim Ber(\sigma(u_{ij})\cdot\sigma(v_{ij}))$. During phase 2 (called "Graph Fitting"), we freeze the previously trained MLP parameters $\theta$ and optimize the mask parameters $\gamma$ using different sets of interventional data $\mathcal{D}_{int(l)}\sim \mathcal{D}_{int}$. To this end, we employ the optimization formulation of \citep{lippe2021efficient}:
    \begin{equation}
        \begin{split}
            \gamma^{*} = (u^{*}, v^{*}) = \argmin_{(u^{*}, v^{*})} &= \mathbb{E}_{l \sim p_{I}(I)} \mathbb{E}_{X \sim \mathcal{D}_{int(l)}} \mathbb{E}_{M  \sim p(M;u,v)}\left[\sum_{i=1}^{N} -log f(X, M_i;\theta_i)\right] \\  & + \underbrace{\lambda_{\text {sparse }} \sum_{i=1}^{N} \sum_{j=1}^{N} \sigma\left(u_{i j}\right) \cdot \sigma\left(v_{i j}\right)}_{:= \text{Regularizer}}
        \end{split}
    \end{equation}
    where $p_{I}(I)$ denotes a distribution over interventions (uniform in our case) and $X \sim \mathcal{D}_{int(l)}$ refers to a a set of data drawn from the interventional dataset $\mathcal{D}_{int(l)}$. As in phase 1 (i.e. distribution fitting), masks $M$ are sampled from $p(M;u,v)$ which represents a distribution over adjacency matrices. For a detailed optimization formulation and gradient derivations, we refer to \citet{lippe2021efficient}.

\clearpage
\subsubsection{Hyperparameter Setup}
\label{appendix:implementation_hyperparameters}

\begin{table}[ht!]
    \centering
    \begin{tabular}{lc}
         \toprule 
         \textbf{Hyperparameters:} & \texttt{Pseudo-LL}, \texttt{EXP-Causal}, \texttt{EXP-AntiCausal}  and \texttt{EXP-Skeleton} \\
         \midrule
         Optimizer:  & Adam \citep{kingma2014adam}\\
         Learning Rate: & $\lbrace 0.1, 0.01, 0.001, \underline{0.0001} \rbrace$ \\
         Weight Decay: & $\lbrace  0.01, 0.001, \underline{0.0} \rbrace$  \\
         Number of iterations: & $\lbrace 500, \underline{1000}, 2000 \rbrace$  \vspace{3mm}\\
         \midrule
         \textbf{Hyperparameters} & \texttt{L-Causal}\\
         \midrule
         Number of Alternating Iterations: & 30 \vspace{1mm}\\
         \textbf{Distribution Fitting:} \\
         Optimizer:  & Adam \citep{kingma2014adam}\\
         Learning Rate: & $\lbrace 0.1, 0.01, 0.001, \underline{0.0001} \rbrace$ \\
         Weight Decay: & $\lbrace  0.01, 0.001, \underline{0.0} \rbrace$  \\
         Number of iterations: & $\lbrace 500, \underline{1000}, 2000 \rbrace$  \vspace{2mm}\\
         \textbf{Graph Fitting$^{*}$:} &  \\
         Optimizer:  & Adam \citep{kingma2014adam}\\
         Learning rate $u$: & $0.005$   \\
         Learning rate $v$: & $0.02$ \\
         Number of iterations: & $100$ \\
         Number of Graphs:     & $100$ \\
         & \vspace{-1mm}\\
         \multicolumn{2}{l}{$^{*}$ All hyperparameters were adopted from \citet{lippe2021efficient}} as we relied on their graph fitting formulation. \vspace{3mm}\\
         \midrule
         \textbf{Hyperparameters} & \texttt{MAML}\\
         \midrule
         \textbf{Inner Loop:} \\
         Optimizer: & SGD \\
         Learning Rate: & $\lbrace \underline{0.1}, 0.01, 0.001 \rbrace$  \\
         Nr. Iterations: & $\lbrace \underline{1}, 2, 5 \rbrace$ \vspace{1mm}\\
         \textbf{Outer Loop:} \\
         Optimizer: & Adam \citep{kingma2014adam}\\
         Learning Rate: & $\lbrace 0.1, 0.01, 0.001, \underline{0.0001} \rbrace$ \\
          Weight Decay: & $\lbrace  0.01, 0.001, \underline{0.0} \rbrace$  \\
         Nr. Iterations: & $\lbrace 500, \underline{1000} \rbrace$ \\
         Nr. Tasks per Iteration: & $\lbrace 10, \underline{20}, 50 \rbrace$ \\
         \bottomrule \\
    \end{tabular}
    \caption{Hyperparameters - Model Training}
    \label{tab:hyperparam_model}
\end{table}

\begin{table}[h!]
    \centering
    \begin{tabular}{cc}
         \toprule
         \textbf{Hyperparameters:} &  Model Adaptation  \\
         Optimizer: & SGD  \\
         Learning Rate: & $\lbrace \underline{0.1}, 0.05, 0.01\rbrace$ \\
         \bottomrule \\
    \end{tabular}
    \caption{Hyperparameters - Model Adaptation}
    \label{tab:hyperparam_adapt}
\end{table}

\clearpage
\subsection{Extended Analysis of Generalization Performance}
In this section, we provide an extended generalization analysis of the one presented in \Cref{sec:generalization}. with additional results and investigations on various settings ($N \in \lbrace 10, 20 \rbrace$, Graphs: \texttt{ER-1}, \texttt{ER-2}, \texttt{ER-3}). We start with an in-depth analysis of a sparse setting  ($N=20$, \texttt{ER-1}, Nr. Training Samples: $1000$) in Section~\ref{appendix:dissection_detail} and thereby highlight the importance of the average evaluation metric dissection. In a second step, we analyze in Section~\ref{appendix:generalization_increasingDim} how the generalization performance is affected as the size of SEM increases. As a final step, we provide the complete results on all sub-metrics across all evaluated settings in Section~\ref{appendix:dissection_all}.

\subsubsection{Case Analysis: N=10, \texttt{ER-1} Graph}
\label{appendix:dissection_detail}
As shown in \Cref{sec:dissection}, an average evaluation metric such as \texttt{NLL-Mean} may not not help us to understand the model's performance in detail and does not provide enough insights where models are prone to fail. In order to highlight the importance of the introduced sub-metrics for evaluating the generalization robustness, we fix a sparse evaluation setting (Graph-Type: \texttt{ER-1}, $N=20$, Nr. Training Samples: $1000$) where all models perform similar with respect to \texttt{NLL-Mean} and dissect the results in detail.

\begin{figure}[h!]
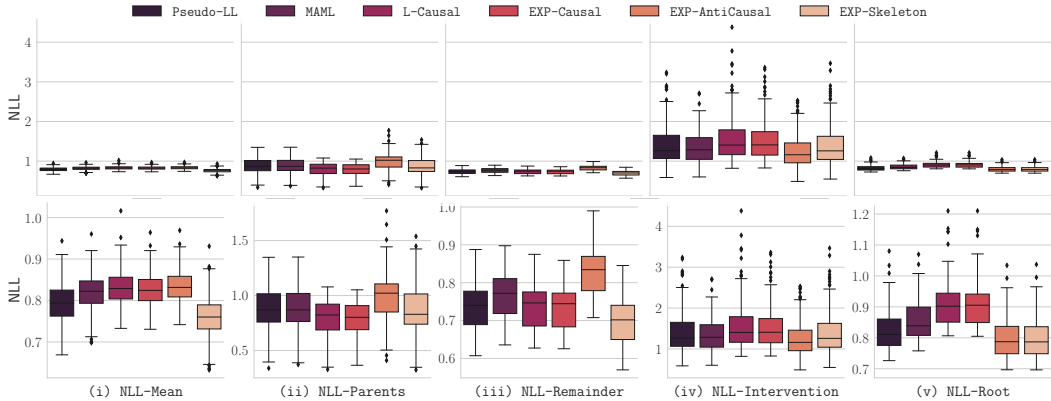

    \centering
    \includegraphics[trim=0 45 0 0, clip, width=1.0\linewidth]{figures/appendix/generalization/dissection_inDetail_ER_1.0_NrVariables_20__shareyTrue_mainPart.pdf}
    \includegraphics[trim=0 0 0 20, clip, width=1.0\linewidth]{figures/appendix/generalization/dissection_inDetail_ER_1.0_NrVariables_20_mainPart.pdf}
    \caption{\textbf{NLL Dissection (Graph-Type: \texttt{ER-1}, $N=20$, Nr. Training Samples: $1000$).} Reporting the sub-metrics on the same scale (top row) clearly shows that \texttt{NLL-Parents} and \texttt{NLL-Intervention} are yielding NLL scores on a different scale. Therefore, we zoom in and show all sub-metrics on their own scale (bottom row). While all models achieve comparable results on most metrics, we observe that \textit{non-causal models} can catastrophically fail to predict the parent variables of an intervened variables (i.e. \texttt{NLL-Parents}). In contrast, \textit{causal models} maintain their performance and outperform all models on the \texttt{NLL-Parents} metric. Furthermore, we observe advantages of \textit{non-causal models} over \textit{causal models} on the \texttt{NLL-Intervention} and \texttt{NLL-Root} metrics which is in line with our expectation as \textit{non-causal models} make use of non-causal predictors.}
    \label{fig:appendix_dissection_detailCase}
\end{figure}

\textbf{Findings.} We observe that \textit{monolithic models} (i.e. \texttt{Pseudo-LL} and \texttt{MAML}) and \texttt{EXP-Skeleton} slightly outperform the two \textit{causal models} (i.e. \texttt{L-Causal}, \texttt{EXP-Causal}) and the \textit{anti-causal model} on the \texttt{NLL-Mean metric} (see \Cref{fig:appendix_dissection_detailCase}). However, by looking at the submetrics in more detail, we find that this performance advantage on \texttt{NLL-Mean} is due to performance differences on \texttt{NLL-Intervention} and \texttt{NLL-Root} where causal models are not expected to achieve similar performance as they only rely on causal predictors (i.e. an empty set for root  and hard-intervened variables). By excluding all variables where the set of causal predictors is empty (i.e. root and intervention nodes), we arrive at the \texttt{NLL-Remainder} metric. On this metric, we observe that all models achieve similar performance ranges except the \textit{anti-causal model} (i.e. \texttt{EXP-AntiCausal}) which yields significantly reduced performance. Finally, we focus on the \texttt{NLL-Parents} metric, where we only evaluate the ability to predict the parents variables $X_{pa(i,G)}$ of a given intervention target $X_i$ which induced the present distribution shift by a perfect intervention $do(X_i)$. While \textit{non-causal models} can catastrophically fail on this task, we observe that \textit{causal models} maintain their performance and outperform all other models. In summary, all models show difficulties to predict the intervened variables as one expects. On all the remaining variables, \textit{causal models} yield the most robust performance without tendencies for catastrophic failure.

\subsubsection{Generalization Performance Across Graphs of Increasing Size}
\label{appendix:generalization_increasingDim}
In this analysis, we seek to investigate how the generalization performance of the considered models changes as the size of the underlying graphs and the corresponding SEM increases. To this end, we fix a sparse class of graphs (i.e. \texttt{ER-1}) and analyze the sub-metrics under training datasets $\mathcal{D}^T$ of different size, i.e. $|\mathcal{D}^T| \in \lbrace 100, 200, 1000 \rbrace$.

\textbf{Findings.} In line with our observation that \textit{structured models} are more sample-efficient than \textit{monolithic models} with respect to the generalization performance, we find that the performance gap between \textit{structured models} and \textit{monolithic models}  widens significantly as the size of the graph increases (i.e. from $N=10$ to $N=20$). In particular, we observe a remarkable generalization behaviour of the causal model \texttt{EXP-Causal} where the true causal structure is provided upfront. Under a fixed size of the training data, the model maintains robust performance over all metrics when the size of the graph increases. Within the models that are not provided with any domain knowledge upfront, we observe that \texttt{L-Causal} and \texttt{MAML} clearly outperform \texttt{Pseudo-LL} on low-sample regimes. As the number of training samples increases, \texttt{L-Causal} is capable of fully identifying the underlying causal structure from data and reaches the same performance as \texttt{EXP-Causal}. In general, the performance gap between the models decreases as the amount of training samples is increased. 

\begin{figure}[h!]
    \centering
    \includegraphics[trim=0 55 0 0, clip, width=1.0\linewidth]{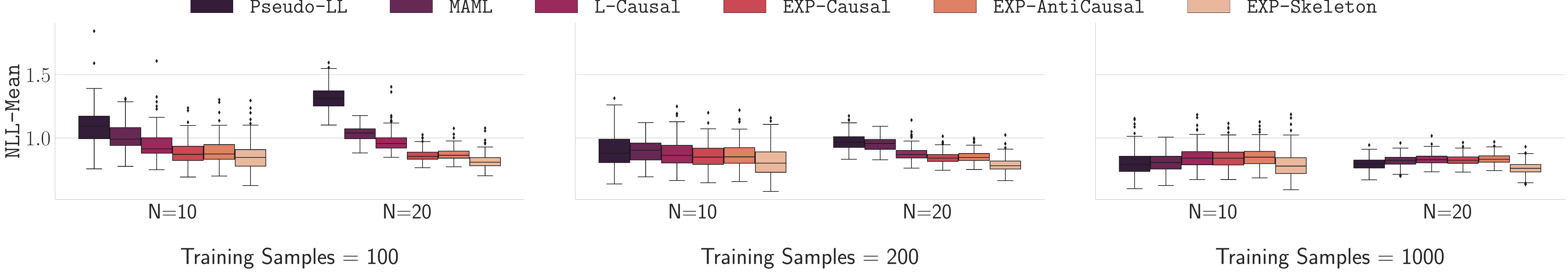}
    \vspace{-6mm}\\
    \begin{tikzpicture}
        \draw [thick,dash pattern={on 7pt off 2pt on 1pt off 3pt}] (-7,0) -- (7,0);
    \end{tikzpicture}
    \includegraphics[trim=0 55 0 35, clip, width=1.0\linewidth]{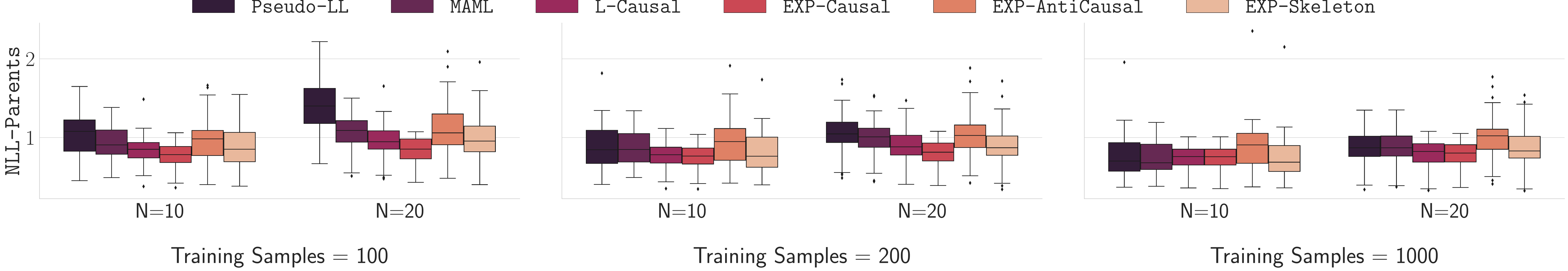}
    \includegraphics[trim=0 55 0 35, clip, width=1.0\linewidth]{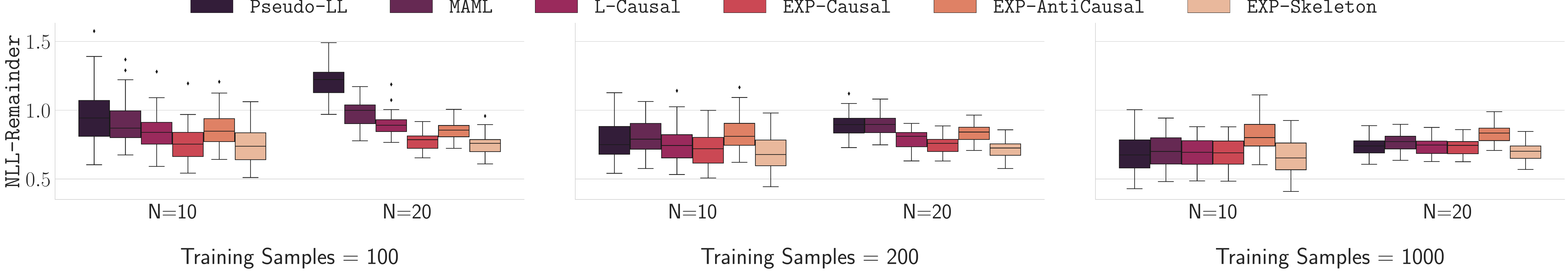}
    \includegraphics[trim=0 55 0 35, clip, width=1.0\linewidth]{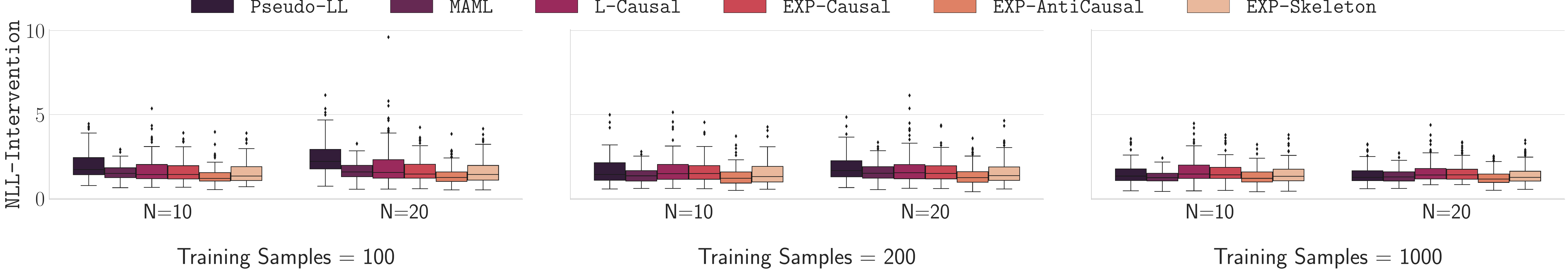}
    \includegraphics[trim=0 0 0 35, clip, width=1.0\linewidth]{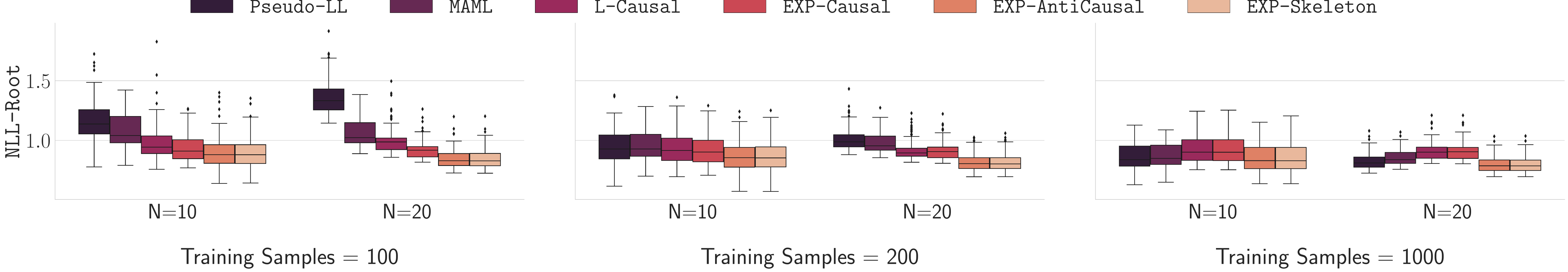}
    \caption{\textbf{Generalization Performance Across Graphs of Increasing Size.} We report the dissected NLL evaluation metrics on \texttt{ER-1} graphs of size $N\in \lbrace 10, 20 \rbrace$. We observe that the performance gap between \textit{structured models} and \textit{monolithic models}  widens significantly as the size of the graph increases.}
    \label{fig:appendix_generalization_overNumberOfVariables}
\end{figure}
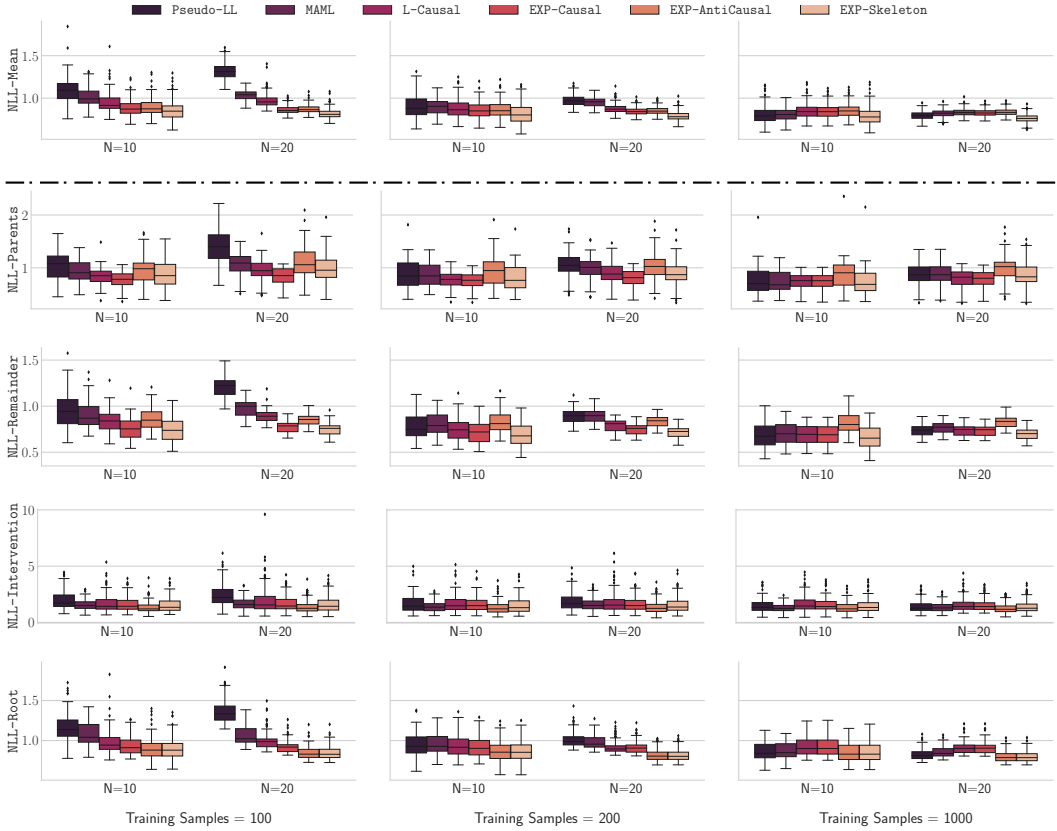

\clearpage
\subsubsection{Dissection - Results Across All Settings}
\label{appendix:dissection_all}
In this section, we report all evaluated (sub)-metrics across \texttt{ER} graphs of varying density (i.e. \texttt{ER-1}, \texttt{ER-2} and \texttt{ER-3} of size $N\in \lbrace 10, 20 \rbrace$ on different amount of training samples $|\mathcal{D}^T| \in \lbrace 100, 200, 400, 1000, 2000 \rbrace$.

\begin{figure}[h!]
    \centering
    \includegraphics[width=1.0\linewidth]{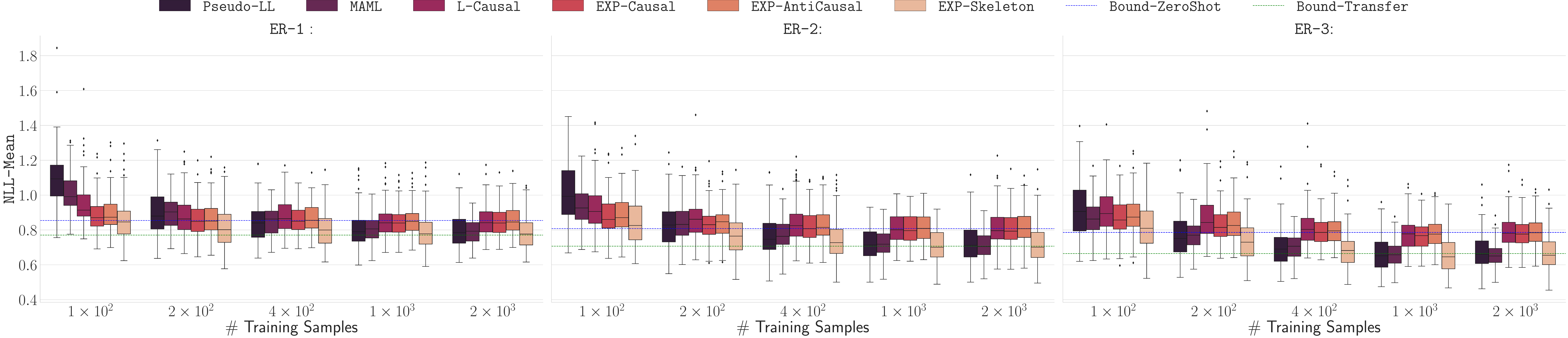}
    \vspace{-6mm}\\
    \begin{tikzpicture}
        \draw [thick,dash pattern={on 7pt off 2pt on 1pt off 3pt}] (-7,0) -- (7,0);
    \end{tikzpicture}
    \includegraphics[trim=0 0 0 108, clip, width=1.0\linewidth]{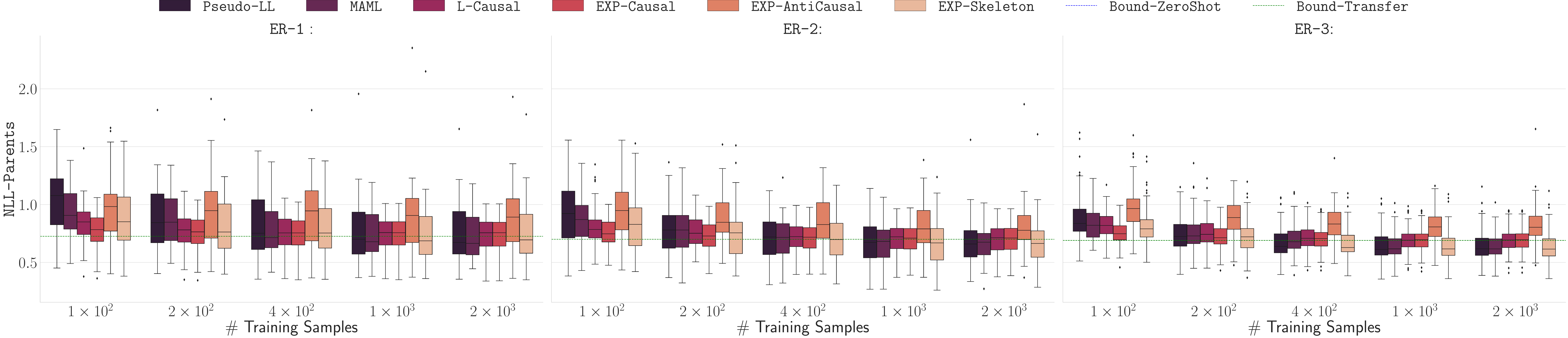}
    \includegraphics[trim=0 0 0 108, clip, width=1.0\linewidth]{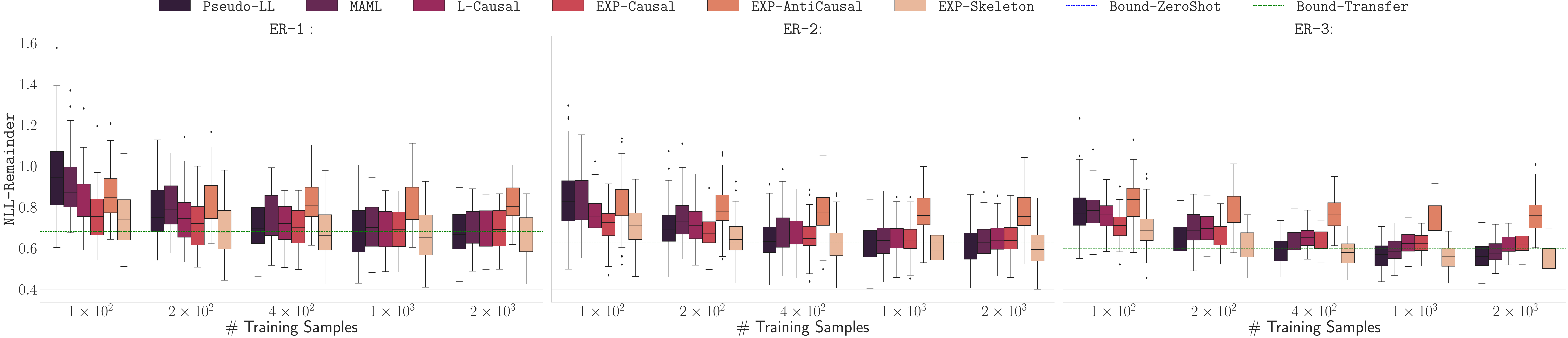}
    \includegraphics[trim=0 0 0 108, clip, width=1.0\linewidth]{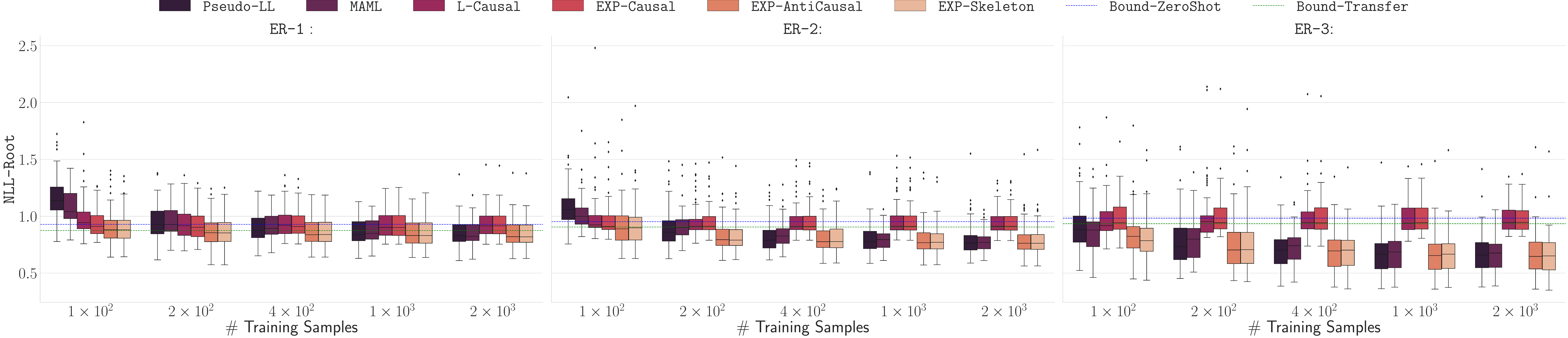}
    \includegraphics[trim=0 0 0 108, clip, width=1.0\linewidth]{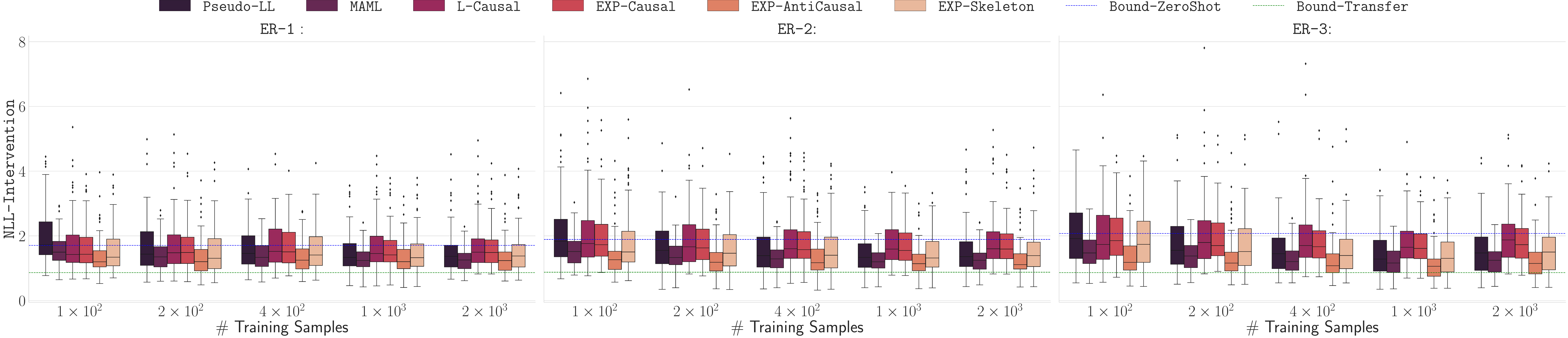}
    \caption{\textbf{Dissection of OOD Generalization with Varying Amounts of Training Data (${\bf N=10}$).} We report all sub-metrics (i.e. one per row) over various \texttt{ER} graphs (i.e. \texttt{ER-1}, \texttt{ER-2} and \texttt{ER-3}). The dissection reveals that  important failure and robustness insights are hidden in the general evaluation score \texttt{NLL-Mean} (top row). While \textit{non-causal models} yield slightly better performance on \texttt{NLL-Root} and \texttt{NLL-Intervention}, we observe that they can catastrophically fail to predict the parent variables of an intervened variables (i.e. \texttt{NLL-Parents}), especially on sparse graphs. Within the \textit{structured models}, we observe that the \texttt{EXP-Skeleton} model that relies on causal and anti-causal predictors performs best across most settings, but is also prone to fail on \texttt{NLL-Parents}. } 
    \label{fig:appendix_dissection_n=10}
\end{figure}

\begin{figure}[h!]
    \centering
    \includegraphics[width=1.0\linewidth]{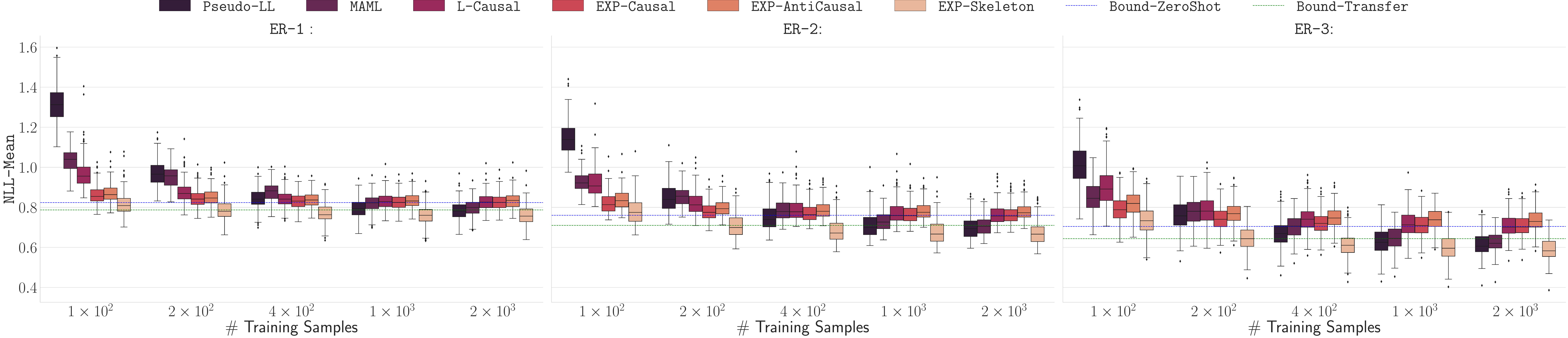}
    \vspace{-6mm}\\
    \begin{tikzpicture}
        \draw [thick,dash pattern={on 7pt off 2pt on 1pt off 3pt}] (-7,0) -- (7,0);
    \end{tikzpicture}
    \vspace{3mm}
    \includegraphics[trim=0 0 0 108, clip, width=1.0\linewidth]{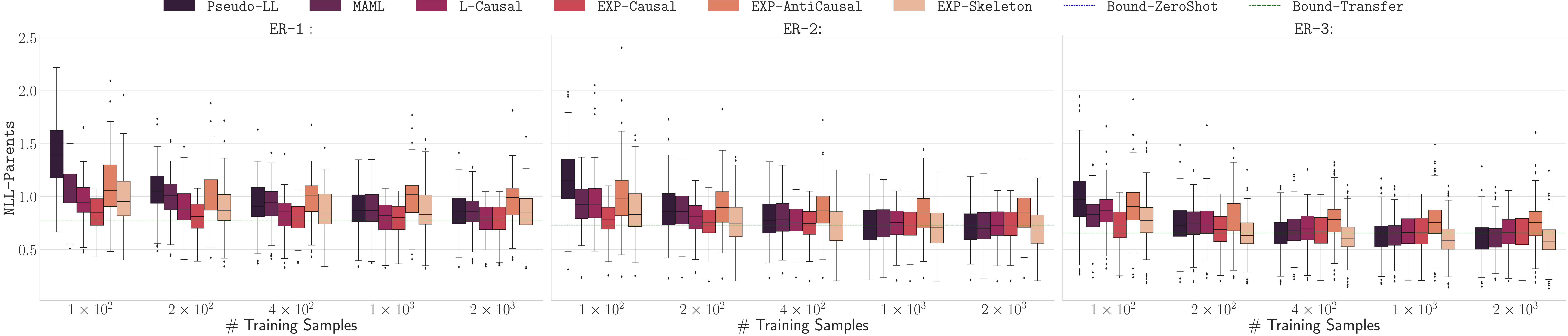}
    \includegraphics[trim=0 0 0 108, clip, width=1.0\linewidth]{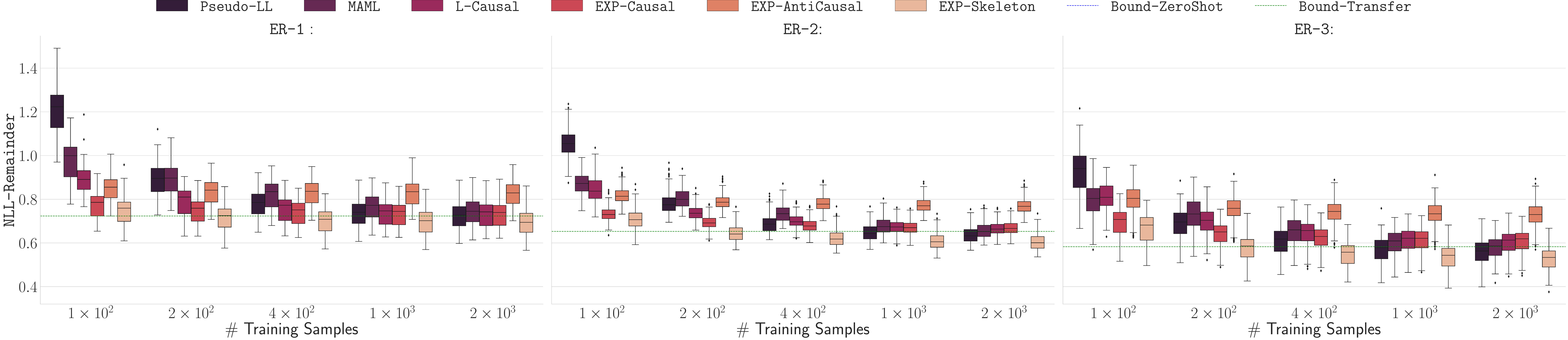}
    \includegraphics[trim=0 0 0 108, clip, width=1.0\linewidth]{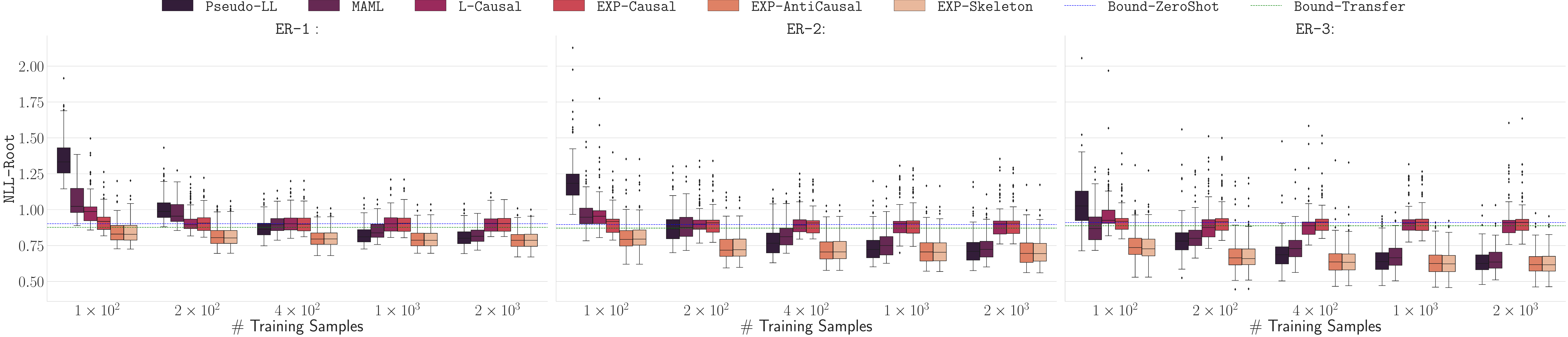}
    \includegraphics[trim=0 0 0 108, clip, width=1.0\linewidth]{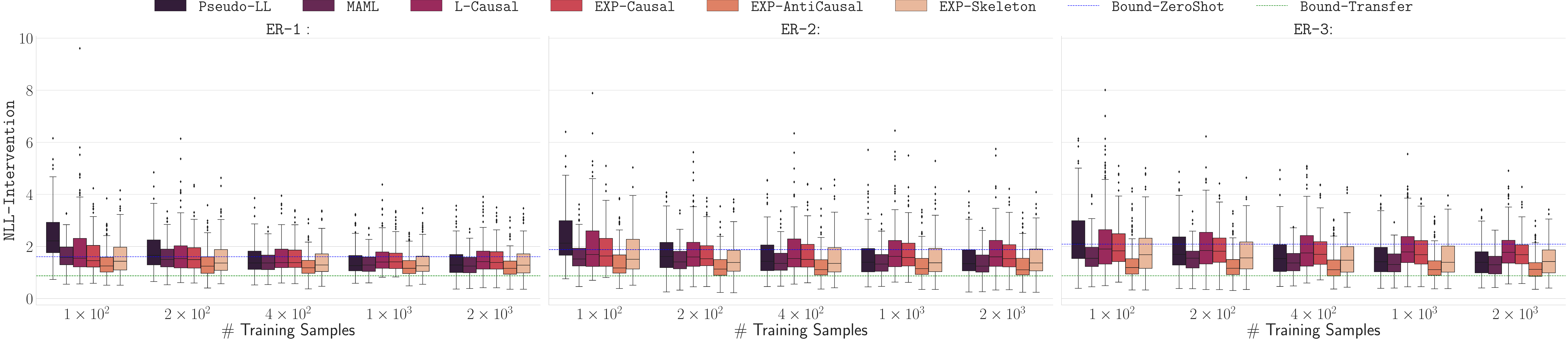}
    \caption{\textbf{Dissection of OOD Generalization with Varying Amounts of Training Data(${\bf N=20}$).} We report all sub-metrics (i.e. one per row) over various \texttt{ER} graphs (i.e. \texttt{ER-1}, \texttt{ER-2} and \texttt{ER-3}). The dissection reveals that  important failure and robustness insights are hidden in the general evaluation score \texttt{NLL-Mean} (top row). While \textit{non-causal models} yield slightly better performance on \texttt{NLL-Root} and \texttt{NLL-Intervention}, we observe that they can catastrophically fail to predict the parent variables of an intervened variables (i.e. \texttt{NLL-Parents}), especially on sparse graphs. Within the \textit{structured models}, we observe that the \texttt{EXP-Skeleton} model that relies on causal and anti-causal predictors performs best across most settings, but is also prone to fail on \texttt{NLL-Parents}. } 
    \label{fig:appendix_dissection_n=20}
\end{figure}

\clearpage
\subsection{Generalization Convergence of Causal Models ( \texttt{L-Causal} vs. \texttt{EXP-Causal})}
In this analsis, we only focus on \textit{causal models} and seek to compare the performance of the \texttt{EXP-Causal} which is provided with the true causal structure upfront, and \texttt{L-Causal} which aims to learn the causal structure from data.

\begin{figure}[h!]
    \centering
    \includegraphics[trim=0 30 0 0, clip, width=1.0\linewidth]{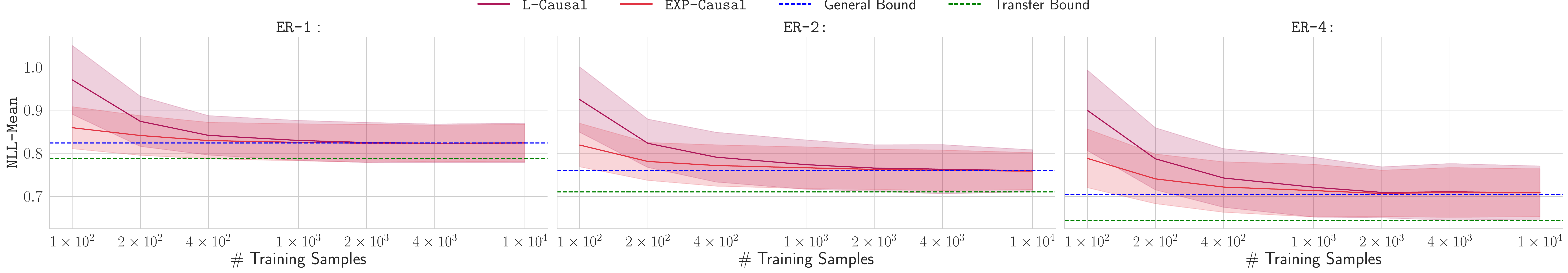}
    \includegraphics[trim=0 0 0 45, clip, width=1.0\linewidth]{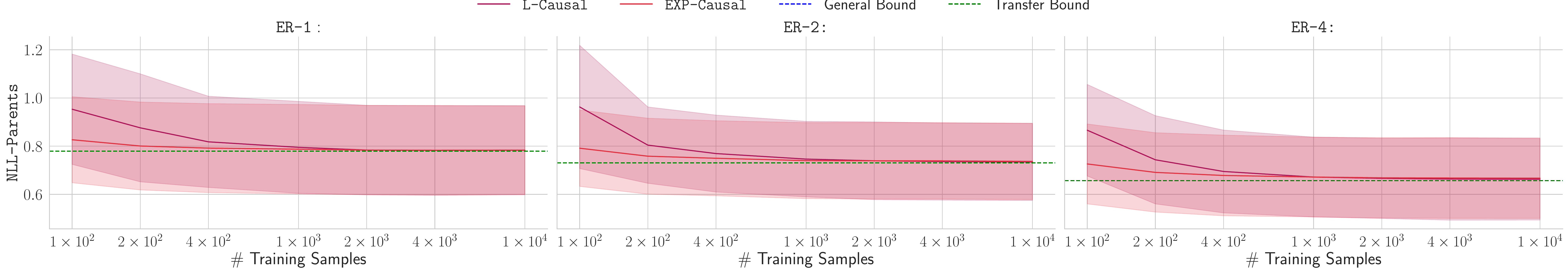}
    \vspace{-6mm}\\
    \begin{tikzpicture}
        \draw [thick,dash pattern={on 7pt off 2pt on 1pt off 3pt}] (-7,0) -- (7,0);
    \end{tikzpicture}
    \vspace{3mm}    
    \includegraphics[width=1.0\linewidth]{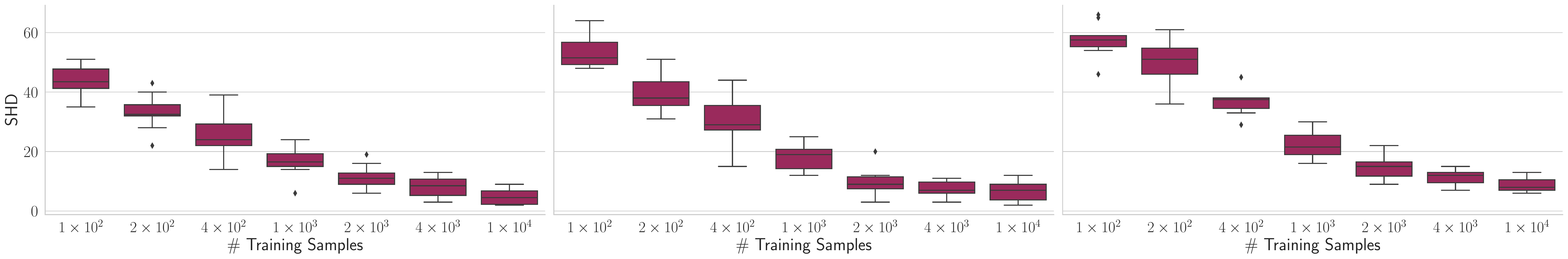}
    \caption{\textbf{Convergence Behaviour of \textit{Causal Models}.} \texttt{EXP-Causal} outperforms \texttt{L-Causal} on low training regimes as \texttt{L-Causal} can only identify the true causal graph with sufficient amounts of samples. With increasing amount of training samples, the structural estimate of \texttt{L-Causal} improves (see bottom row) and hence the generalization performance improves and converges to \texttt{Bound-ZeroShot}.}
    \label{fig:appendix_causalModels_convergence}
\end{figure}
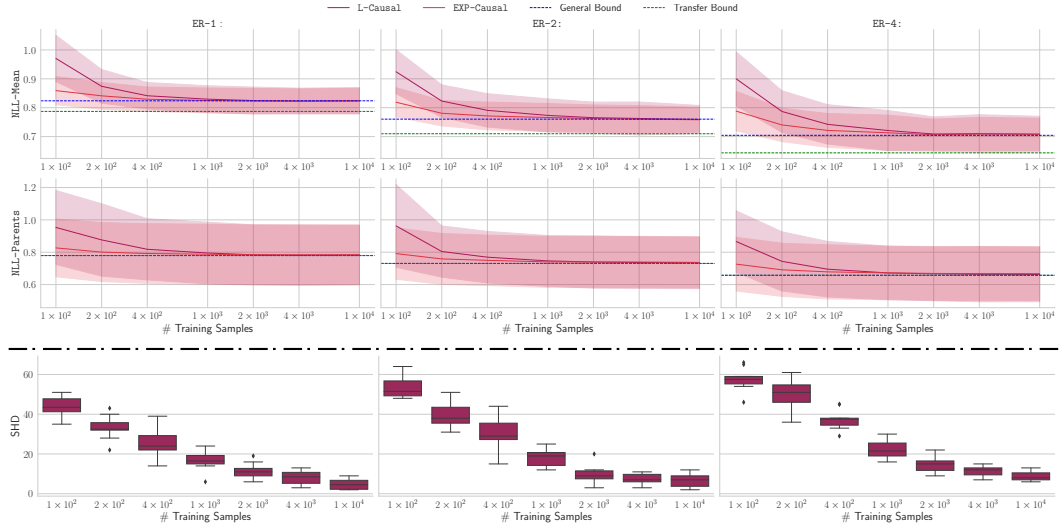

\textbf{Findings.} We observe that \texttt{EXP-Causal} outperforms \texttt{L-Causal} on low training regimes as the employed causal discovery framework can only identify the true causal graph with sufficient amounts of samples (see \Cref{fig:appendix_causalModels_convergence}). With increasing amounts of samples, the learned causal structure of \texttt{L-Causal} gets closer to the ground-truth structure (see bottom row of \Cref{fig:appendix_causalModels_convergence} for Structural Hamming Distance (SHD) between learned and true structure) and hence the generalization performance improves. Both models attain \texttt{Bound-ZeroShot} (blue) as expected with sufficient amount of samples. In addition, we observe slower convergence of \texttt{L-Causal} to \texttt{Bound-ZeroShot} on dense graphs than on sparse graph, as the identification of the causal structure is more challenging in such settings \citep{ke2019learning, scherrer2021learning}.  

\clearpage
\subsection{Extended Analysis of Adaptation Performance}
In this section, we expand our few-shot adaptation analysis from \Cref{sec:adaptation} with respect to speed of adaptation and overfitting behaviour in Section \ref{appendix:adaptation_performance}, and the effect on the parameter space $\theta$ in Section \ref{appendix:parameter_space_analysis}. In addition, we provide further results and analysis on the regularized adaptation objective.

\subsubsection{Adaptation Performance}
\label{appendix:adaptation_performance}
As in \Cref{sec:adaptation}, we fix the training data size of all models to be $10^3$ samples, as all models have converged on the generalization performance by then.  We analyze the speed of adaption of different models by evaluating their adaption performance when fine-tuning using different amounts of adaptation data.

\begin{figure}[h!]
    \centering
    \includegraphics[width=1.0\linewidth]{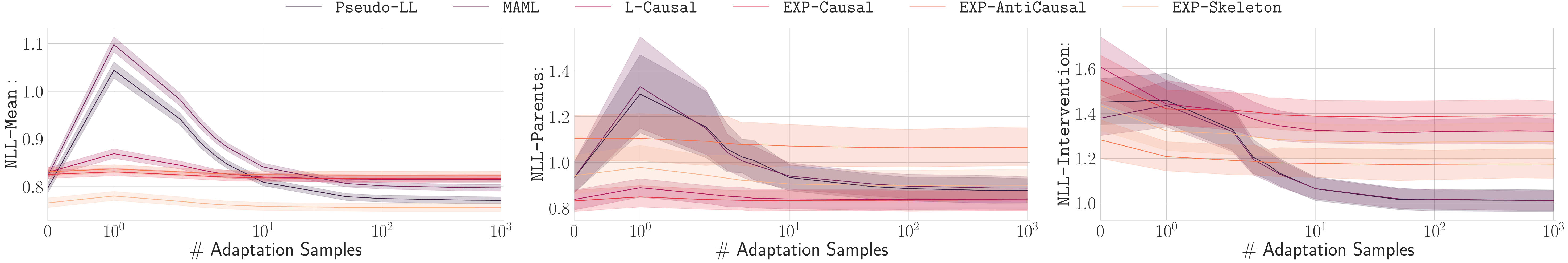}
    (i) Graphs: \texttt{ER-1} \vspace{2mm}\\
    \includegraphics[trim=0 0 0 30, clip, width=1.0\linewidth]{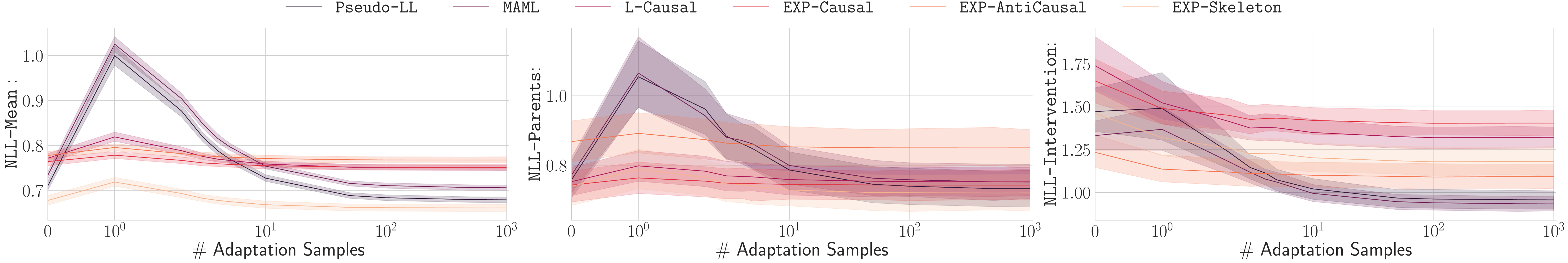}
    (ii) Graphs: \texttt{ER-2} \vspace{2mm}\\
    \includegraphics[trim=0 0 0 30, clip, width=1.0\linewidth]{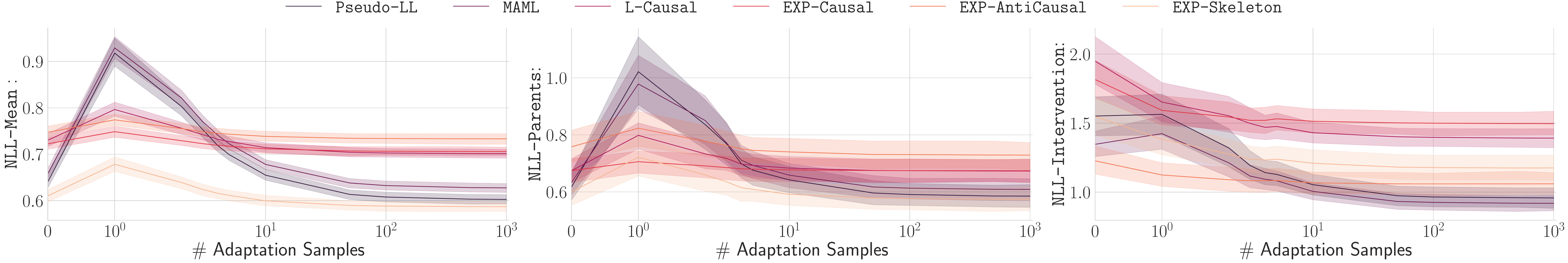}
    (iii) Graphs: \texttt{ER-3} \vspace{2mm}\\
    \caption{\textbf{Speed of Adaptation in terms of different metrics ($N=20, \mathcal{D}^T=1000$, $3$  Gradient Steps).} \textit{Structured models} adapt considerably faster than \textit{monolithic models} across all settings and metrics. \textit{Monolithic models} show a sensitivity to overfitting on all classes of graph if only low amounts of adaptation samples are available. In contrast, \textit{structured models} adapt smoothly to the transfer distribution with significantly reduced overfitting effects.}
    \label{fig:appendix_analysis_adaptation}
\end{figure}

\textbf{Findings.}  Across all evaluated classes of graphs (e.g. \texttt{ER-1}, \texttt{ER-2} and \texttt{ER-3}), we observe that \textit{structured models} adapt considerable faster than \textit{monolithic models} with respect to the required amount of adaptation samples. When doing a few gradient steps using SGD (i.e. 3 steps with a learning rate of $0.1$), we already observe strong overfitting effects for the considered \textit{monolithic models} on all evaluated metrics except \texttt{NLL-Intervention} (see  \Cref{fig:appendix_analysis_adaptation}). By inspecting the \texttt{NLL-Parents} metric, we observe the robustness of the two causal models (i.e. \texttt{L-Causal} and \texttt{EXP-Causal}), especially on sparse graphs. 

\clearpage
\subsubsection{Parameter Space Analysis}
\label{appendix:parameter_space_analysis}
Keeping the adaptation performance of the previous section in mind, we now expand our analysis on the parameter space $\theta$. We seek to answer if the adaptation performance is related to the changes in parameter space $\theta$.

\textbf{Findings.} We find that the overfitting behaviour of \textit{monolithic models} is correlated with the observed updates in parameter space. For the range of adaptation samples where the \textit{monolithic models} are prone to overfit (i.e. $1$ to $10$ adaptation samples), we observe high gradient magnitudes on the non-intervened modules (referred to as other  modules) in \textit{monolithic models} compared to the relatively small updates of \textit{structured models}. As the size of adaptation samples increases (i.e. 100 adaptation samples), we observe significantly reduced gradient magnitudes on non-intervened modules and lower overfitting effects. Within the \textit{structured models} that are built upon structural domain knowledge, we observe that \texttt{EXP-Causal} and \textit{EXP-AntiCausal} yield relatively small gradient updates compared to \texttt{EXP-Skeleton}. In addition, we observe that the \texttt{anti-causal model} yields lower updates on the intervened module as expected as it relies on anti-causal predictors of the intervened variable children, that were not affected by the intervention.

\begin{figure}[h!]
    \centering
    \includegraphics[width=1.0\linewidth]{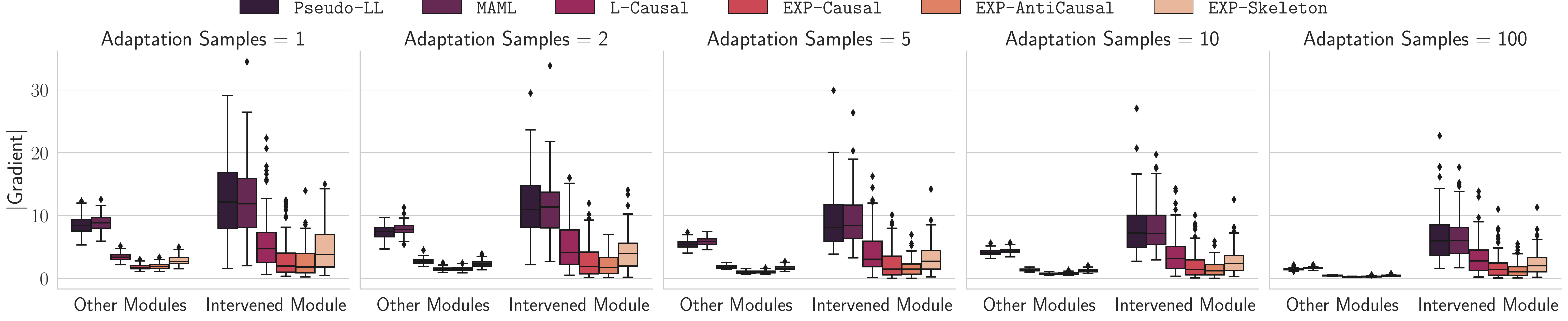}
    (i) Graphs: \texttt{ER-1} \\
    \includegraphics[trim=0 0 0 25, clip, width=1.0\linewidth]{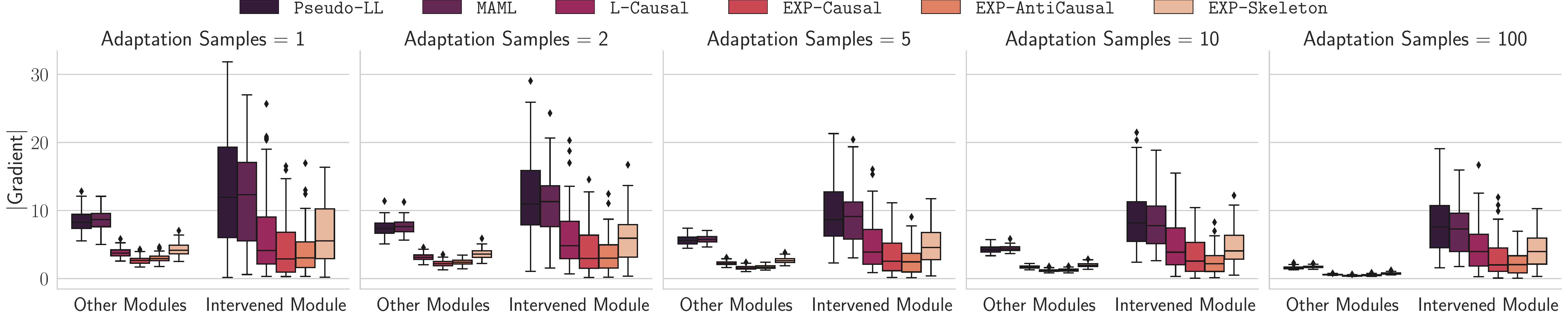}
     (ii) Graphs: \texttt{ER-2} \\
    \includegraphics[trim=0 0 0 25, clip, width=1.0\linewidth]{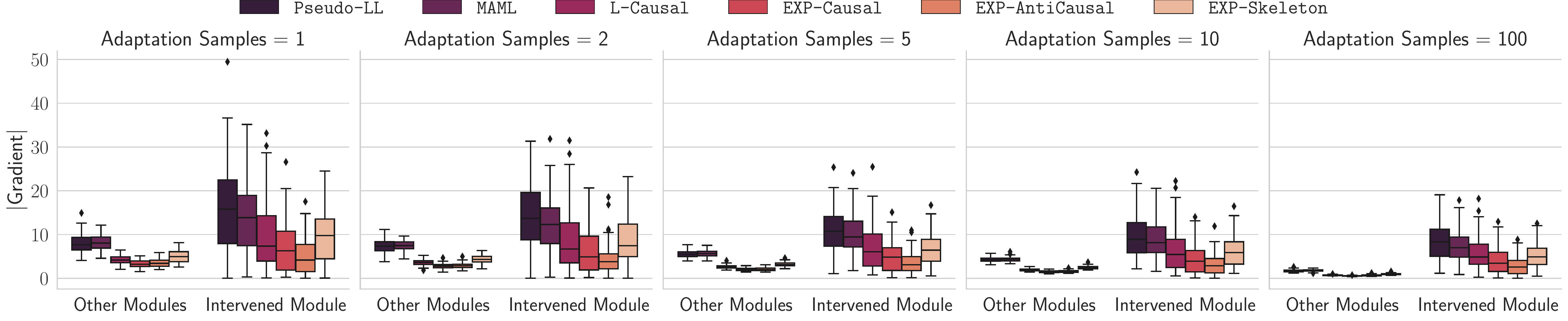}
     (iii) Graphs: \texttt{ER-3} \\
    \caption{\textbf{Parameter Space Analysis ($N=20, \mathcal{D}^T=1000$).} While adapting to a shift in distribution with an unconstrained adaptation objective using a single gradient step, \textit{monolithic models} update most modules that were not affected by the intervention quite heavily compared to \textit{structured models}. \textit{Causal and anti-causal models} show remarkable adaptation behaviour in parameter space with localized updates on intervened modules.}
    \label{fig:my_label}
\end{figure}

\clearpage
\subsubsection{Regularized Adaptation}
\begin{figure}[h!]
    \centering
    \includegraphics[width=1.0\linewidth]{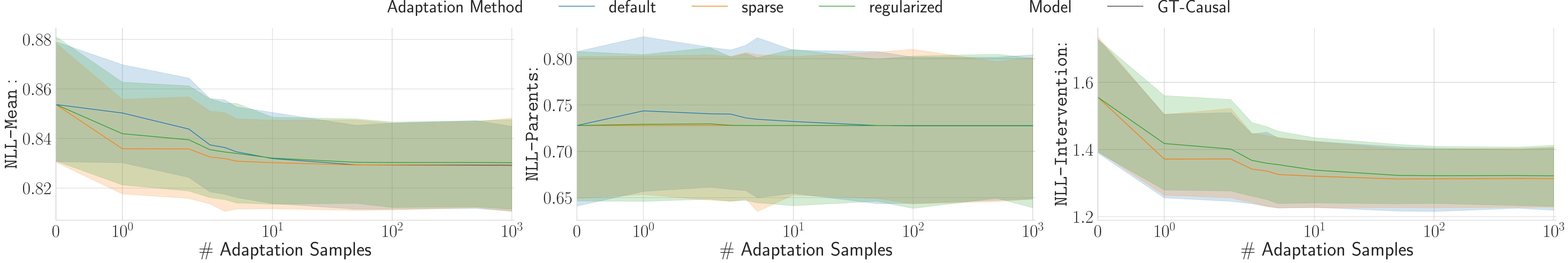}
    (i) Gradient Steps: $5$
    \includegraphics[trim=0 0 0 30, clip, width=1.0\linewidth]{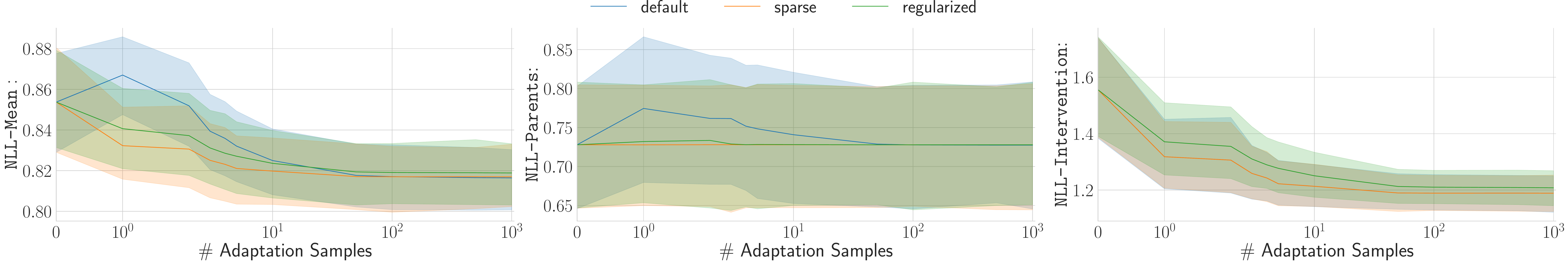}
    (ii) Gradient Steps: $10$
    \caption{\textbf{Regularized Adaptation: Effects on Speed of Adaptation ($N=10, \mathcal{D}^T=1000$).} For $5$ gradients steps using SGD with a learning rate of $0.1$, we observe continuously improving adaptation with respect to the \texttt{NLL-Mean} metric on all regularization techniques. With $10$ gradient steps, we observe an overfitting behaviour of the unconstrained adaptation objective if only low amounts of adaptation samples are available. In contrast, the sparse and regularized adaptation objective still yield continuous improvements, even if only low few adaptation samples are available. It is noteable, that the regularized adaptation objective yields nearly the same performance as the sparse adaptation objective, even though the sparse adaptation objective leverages a supervised signal (i.e. knowledge of the intervention location) in the present setting. }
    \label{fig:appendix_regularized_speed}
\end{figure}

\begin{figure}[h!]
    \centering
    \includegraphics[width=1.0\linewidth]{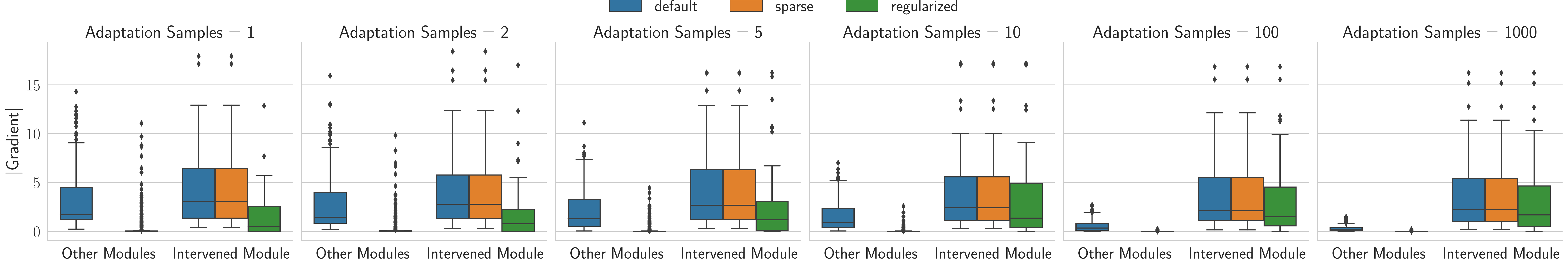}
    \caption{\textbf{Regularized Adaptation: Effects on Parameter Space $\theta$ ($N=10, \mathcal{D}^T=1000$).} With respect to the parameter space, we observe that the regularized adaptation objective yields smaller updates on the intervened module if only low amounts of adaptation samples are available. In general, the regularized adaptation objective is capable of identifying the intervened mechanisms and only performs updates of low gradient-magnitude on non-intervened modules whereas the unconstrained adaptation objective yields updates of greater magnitude. As the number of adaptation samples increases, the regularized objective yields similar updates on the intervened mechanisms as the other two objectives.  }
    \label{fig:appendix_regularized_parameter}
\end{figure}

In this section, we provide further insights on the effects of a a regularized adaptation objective on top of \texttt{EXP-Causal}. We report speed of adaptation in \Cref{fig:appendix_regularized_speed} and analyses on the parameter space in \Cref{fig:appendix_regularized_parameter}.

\textbf{Findings.} We observe that the regularized adaptation objective improves the adaptation performance on low amounts of adaptation samples considerably. Our results indicate that the adaptation objective prevents from overfitting if multiple gradient steps are performed. It is noteable, that the regularized adaptation objective yields nearly the same performance as the sparse adaptation objective, even though the sparse adaptation objective leverages a supervised signal (i.e. knowledge of the intervention location) in the present setting.

\end{document}